\documentclass[11pt,a4paper]{article}
\usepackage[square,sort,comma,numbers]{natbib}
\usepackage{amsmath,amsfonts}
\usepackage{graphicx}
\usepackage[colorlinks=true]{hyperref}
\usepackage{amsfonts}
\usepackage{color}
\usepackage{epsfig}
\usepackage{amssymb}
\usepackage{textgreek}
\newcommand{\argmin}{\mathrm{argmin}}
\newcommand{\sparse}{\mathrm{sparse}}
\newcommand{\Prox}{\mathrm{Prox}}
\newcommand{\wh}[1]{{\widehat{#1}}}
\def\Argmin{\mathop\mathrm{Argmin}}
\def\la{{\langle}}
\def\ra{{\rangle}}
\def\lowkap{{\underline{\kappa}}}

\def\risk{{\hbox{\rm Risk}}}

\def\rank{{\hbox{\rm rank}}}
\def\card{{\mathrm{card}}}

\def\cP{{\cal P}}
\def\Diag{{\hbox{\rm Diag}}}
\def\bR{{\mathbf{R}}}
\def\bZ{{\mathbf{Z}}}

\def\L{{\cal L}}

\def\prob1{$(SO)$}
\def\SG{\mathrm{Sub}{\cal G}}

\def\P{{\cal P}}

\def\M{{\cal M}}
\def\X{{\cal X}}

\def\N{{\cal N}}

\def\bE{{\mathbf{E}}}

\def\Prob{\hbox{\rm Prob}}
\DeclareMathOperator*{\med}{median}

\def\Row{\hbox{\rm Row}}

\def\qed{\hfill$\Box$}
\def\Card{\mathop{\hbox{\rm Card}}}

\def\Tr{{\hbox{\rm Tr}}}

\newcommand{\be}{\begin{eqnarray}}
\newcommand{\ee}[1]{\label{#1}\end{eqnarray}}
\newcommand{\nn}{\nonumber \\}

\newcommand{\D}{{\cal D}}

\newcommand{\four}{ \mbox{\small$\frac{1}{4}$}}

\newcommand{\rf}[1]{(\ref{#1})}
\newcommand{\half}{ \mbox{\small$\frac{1}{2}$}}

\newcommand{\bse}{\begin{eqnarray*}}
\newcommand{\ese}{\end{eqnarray*}}

\definecolor{MyDarkBlue}{rgb}{0,0.08,0.45}
\definecolor{MyViolet}{rgb}{0.45,0.08,0.95}
\definecolor{MyBrown}{rgb}{0.45,0.08,0}

\newcommand{\hide}[1]{{}}

\newtheorem{theorem}{Theorem}[section]

\newtheorem{lemma}{Lemma}[section]
\newtheorem{proposition}{Proposition}[section]

\def\ul{{\underline{\ell}}}
\def\ol{{\overline{\ell}}}
\makeatletter
\renewcommand*{\@fnsymbol}[1]{\ifcase#1\or*\else\@arabic{\numexpr#1-1\relax}\fi}
\makeatother
\newcommand*\samethanks[1][\value{footnote}]{\footnotemark[#1]}

\def\myu{\mathfrak{u}}
\def\myU{\mathfrak{v}}
\def\mynu{\mbox{\textnu}}

\begin{document}
\title{Sparse recovery by reduced variance stochastic approximation}

\author{Anatoli Juditsky\thanks{LJK, Universit\'e Grenoble Alpes, 700 Avenue Centrale,  38401 Domaine Universitaire de Saint-Martin-d'Hères, France,
{\tt anatoli.juditsky@univ-grenoble-alpes.fr}}~$^,$\thanks{Research of this
author was supported by MIAI {@} Grenoble Alpes (ANR-19-P3IA-0003).}
\and Andrei Kulunchakov\thanks{
Universit\'e Grenoble Alpes, Inria, CNRS, Grenoble INP, LJK, Grenoble, 38000, France, {\tt andrei.kulunchakov@inria.fr}}
\and Hlib Tsyntseus\thanks{LJK, Universit\'e Grenoble Alpes, 700 Avenue Centrale,  38401 Domaine Universitaire de Saint-Martin-d'Hères, France, {\tt hlib.tsyntseus@univ-grenoble-alpes.fr}}~$^,$\samethanks[2]
}
\maketitle

\begin{abstract}
{In this paper, we discuss application of iterative Stochastic Optimization routines to the problem of sparse signal recovery
from noisy observation.
Using Stochastic Mirror Descent algorithm as a building block, we develop a multistage procedure for recovery of sparse solutions to Stochastic Optimization problem under assumption of smoothness and quadratic minoration on the expected objective. An interesting feature of the proposed algorithm is linear convergence of the approximate solution during the preliminary phase of the routine when the component of stochastic error in the gradient observation which is due to bad initial approximation of the optimal solution is larger than the ``ideal'' asymptotic error component owing to observation  noise ``at the optimal solution.''  We also show how one can straightforwardly enhance reliability of the corresponding solution {by} using Median-of-Means like techniques.\par
We illustrate the performance of the proposed algorithms in application to classical problems of recovery of sparse and low rank signals in the generalized linear regression framework. We show, under rather weak assumption on the regressor and noise distributions, how they lead to parameter estimates which obey (up to factors which are logarithmic in problem dimension and confidence level) the best known to us accuracy bounds.}
\\
{\bf 2000 Math Subject Classification:} 62G08, 62G35, 62J07,  90C15
\end{abstract}

\section{Introduction}
In this paper, we consider the Stochastic Optimization problem of the form
\be
g_*=\min_{x\in X} \big\{g(x)=\bE\{G(x,\omega)\}\big\}
\ee{prob0st}
where $X$ is a given convex and closed subset of a Euclidean space $E$, $G:\,X\times \Omega\to \bR$ is a smooth convex mapping, and $\bE$ stands for the expectation with respect to unknown distribution of $\omega\in \Omega$ (we assume that the corresponding expectation exists for every $x\in X$). As it is usual in this situation, we suppose that we have access to a stochastic ``oracle''  supplying ``randomized'' information about $g$; we assume that the problem is solvable with the optimal solution $x_*$ which is {\em sparse} (we consider a general notion of sparsity structure of $x_*$ as defined in Section \ref{sec:probs} which comprises ``usual'' sparsity,  group sparsity, and low rank matrix structures as basic examples).

Our interest in \rf{prob0st} is clearly motivated by statistical applications.
Recently,  different techniques of estimation and selection under sparsity and low rank constraints gained a lot of attention, in particular, in relation with the sparse linear regression problem in which unknown $s$-sparse (i.e., with at most $s$ nonvanishing components) vector $x_*\in \bR^n$ of regression coefficients is to be recovered from the linear noisy observation
\be
\eta=\Phi^T x_*+\sigma\xi,
\ee{lreg0} where $\Phi\in \bR^{n\times N}$ is the regression matrix, and $\xi\in \bR^N$ is zero-mean noise with unit covariance matrix; we are typically interested in the situation where the problem dimension is large, i.e. when $n\gg N$.
Note that the problem of sparse recovery from observation \rf{lreg0} with random regressors (columns of the regression matrix $\Phi$) $\phi_i,\,i=1,...,N$  can be cast as Stochastic Optimization. For instance, assuming that regressors $\phi_i$ and noises $\xi_i$, $i=1,..., N$, are identically distributed, we may consider Stochastic Optimization problem
\be
\min_{x\in X} \Big\{g(x)=\half \bE\{(\eta_1-\phi^T_1x)^2\}\Big\}
\ee{sa0}
over $s$-sparse $x\in X$.  There are essentially two approaches to solving \rf{sa0}.
Note that observations $\eta_i$ and $\phi_i$ provide us with unbiased estimates $G(x,\omega_i=[\phi_i,\eta_i])=\half \|\eta_i-\phi^T_ix\|_2^2$ of the problem objective $g(x)$. Therefore, one can build a Sample Average Approximation (SAA) \[
\wh g(x)={1\over N}\sum_{i=1}^N G(x,\omega_i)=\mbox{\small $1\over 2N$} \|\eta-\Phi^Tx\|_2^2\] of the objective $g(x)$ of \rf{sa0} and then solve the resulting Least Squares problem by a deterministic optimization routine.
A now standard approach to enhancing the sparsity of solutions is to use iterative thresholding~\cite{blumensath2009iterative,jain2014iterative,Barber1,Barber2}.
When applied to the linear regression problem \rf{sa0}, this technique amounts to using a gradient descent to minimize the Least Squares objective $\wh g$ in combination with thresholding of approximate solutions to enforce sparsity. Another approach which refers to $\ell_1$- and nuclear norm minimization allows to reduce problems of sparse or low rank recovery to convex optimization. In particular, sparse recovery by Lasso and Dantzig Selector has been extensively studied in the statistical literature
\cite{candes2006stable,candes2007dantzig,bickel2009simultaneous,van2009conditions,candes2009near,candes2011probabilistic,raskutti2010restricted,fazel2008compressed,candes2009exact,
candes2011tight,juditsky2011accuracy,negahban2011estimation,koltchinskii2011nuclear,rudelson2012reconstruction,dalalyan2019outlier}, among others). For instance, the celebrated Lasso estimate $\wh x_{N,\mathrm{lasso}}$ in the sparse linear regression problem is a solution to the $\ell_1$-penalized Least Squares problem
\be
\wh x_{N,\mathrm{lasso}}\in \Argmin_{x}\left\{\mbox{\small $1\over 2N$}\|\eta-\Phi^Tx\|_2^2+\lambda\|x\|_1
\right\}
\ee{lasso}
where $\lambda\geq 0$ is the algorithm parameter. Several conditions which ensure recovery with ``small error'' of any sparse or low rank signal using $\ell_1$- and nuclear norm minimization are proposed. In particular, recovery of {\em any} $s$-sparse  (i.e., with at most $s$ nonvanishing components) vector $x_*$ is possible with ``small error'' if the empirical regressor covariance matrix $\wh \Sigma={1\over N}\Phi\Phi^T$ verifies a certain restricted conditioning assumption, e.g., Restricted Eigenvalue (RE) \cite{bickel2009simultaneous} or Compatibility condition \cite{van2009conditions}. The latter conditions very roughly mean that
for all vectors  $z$ which are ``approximately sparse,'' i.e., which are close to vectors with only $s$ nonvanishing entries, $\|\wh \Sigma z\|_2\geq \lambda \|z\|_2$. The good news is that although these conditions are typically difficult to verify for individual matrices $\Phi$, they are satisfied for several families of random matrices, such as Rademacher (with independent random $\pm 1$ entries) and Gaussian matrices, matrices uniformly sampled from Fourier or Hadamard bases of $\bR^n$, etc.
 For instance, when columns $\phi_i$ of $\Phi$ are sampled independently from normal distribution $\phi_i\sim \N(0,\Sigma)$ with covariance matrix $\Sigma$ with bounded diagonal elements which satisfies $\kappa_{\Sigma} I\preceq\Sigma$ (here $I$ is the $n\times n$-identity matrix),\footnote{Here and in the sequel, we use notation $A\preceq B$ for $n\times n$ symmetric matrices $A$ and $B$ such that $B-A\succeq0$, i.e. $B-A$ is positive semidefinite.} $\kappa_\Sigma>0$, RE condition holds with high probability for $s$ as large as $O\left({N\kappa_\Sigma\over \ln[n]}\right)$ \cite{raskutti2010restricted}.\footnote{The reader acquainted with the compressive sensing theory will notice that the setting of the $\ell_1$-recovery problem considered in this paper is different from the s``tandard setting,''
 but is rather similar in spirit to that in \cite{candes2009near,candes2011probabilistic,adcock2017breaking,bigot2016analysis,boyer2019compressed}. Although, unlike
 \cite{adcock2017breaking,bigot2016analysis,boyer2019compressed} we do not assume any special structure of $x_*$ apart from its sparsity, we suppose random regressors to be independent of $x_*$, while in the ``standard setting'' one allows for the ``worst case $x_*$'' which may depend on the particular realization of the matrix of regressors. Nevertheless, we do not know any result stating that a recovery in the present setting is possible under ``essentially less restrictive'' assumptions than those for the ``standard'' $\ell_1$ recovery.}
%
%
%
\par
The Restricted Strong Convexity (RSC) condition, analogous to the RE or Compatibility condition also ensure
that iterative thresholding procedures converge linearly to an approximate solution with accuracy which is similar to that of Lasso or Dantzig Selector estimation \cite{Barber1,Barber2} in this case.
%
  \par
Another approach to solving \rf{prob0st} which refers to Stochastic Approximation (SA) may be used whenever there is a ``stochastic oracle'' providing an unbiased stochastic observation of the gradient $\nabla g$ of the objective $g$ of \rf{prob0st}. For instance, note that the observable quantity $\nabla G(x,\omega_i)=\phi_i(\phi^T_ix-\eta_i)$ is an unbiased estimate of the gradient $\nabla g(x)$ of the objective of \rf{sa0}, and so an iterative algorithm of Stochastic Approximation type can be used to build approximate solutions to \rf{sa0}.
In particular, different versions of Stochastic Approximation procedure were applied to solve \rf{sa0} under $\ell_1$
and sparsity constraint. Recall, that we are interested in high-dimensional problems, we are looking for bounds for recovery error which are ``essentially independent'' (logarithmic, at most)
in problem dimension $n$. This requirement rules out the use of standard ``Euclidean'' Stochastic Approximation. Indeed, typical bounds for the expected inaccuracy $\bE\{g(\wh x_N)\}-g_*$ of
Stochastic Approximation contains the term proportional to $\sigma^2 \bE\{\|\phi_1\|_2^2\}$ and thus proportional to $n$ in the case of ``dense'' regressors with $\bE\{\|\phi_1\|_2^2\}=O(n)$.
Therefore, unless regressors $\phi$ are sparse (or possess a special structure, e.g., when $\phi_i$ are low rank matrices in the case of low rank matrix recovery), standard
Stochastic Approximation leads to accuracy bounds for sparse recovery which are proportional to dimension $n$ of the parameter vector \cite{nguyen2017linear}.
In other words, our application calls for non-Euclidean Stochastic Approximation procedures, such as Stochastic Mirror Descent algorithm \cite{nemirovskii1979complexity}.

In particular, \cite{shalev2011stochastic,srebro2010smoothness} study the properties of Stochastic Mirror Descent algorithm under sub-Gaussian noise assumption and show that approximate solution $\wh x_{N}$ after
 $N$ iterations of the method attains the bound $g(\wh x_N)-g_*=O\left({\sigma \sqrt{s\ln (n)/N}}\right)$, often referred to as ``slow rate'' of sparse recovery. In order to improve the error estimates of Stochastic Approximation one may use multistage algorithm under strong or uniform convexity assumption \cite{juditsky2011optimization,juditsky2014deterministic,ghadimi2013optimal}. However, such assumptions do not hold in the problems such as sparse linear regression problem,\footnote{More generally, strong convexity of the objective associated with smoothness is a feature of the Euclidean setup. For instance, the conditioning of a smooth objective (the ratio of the Lipschitz constant of the gradient to the constant of strong convexity) when measured with respect to the $\ell_1$-norm cannot be less than $n$ (the problem dimension) \cite{juditsky2014deterministic}.} where they are replaced by Restricted Strong Convexity conditions. For instance, the authors of \cite{agarwal2012stochastic,gaillard2017sparse}
develop a multistage procedure targeted at sparse recovery stochastic optimization problem \rf{prob0st} based on SMD algorithm of  \cite{juditsky2006generalization,nesterov2009primal} under bounded regressor and sub-Gaussian noise  assumption. They show, for instance, that when applied to the sparse linear regression, the $\ell_2$-error $\|\wh{x}_N-x_*\|_2$ of the approximate solution $\wh x_N$ after $N$ iterations of the proposed routine converges at the rate $O\left({\sigma\over \kappa_\Sigma}\sqrt{s\ln n\over  N}\right)$
with high probability. While this ``asymptotic'' rate coincides with the best rate attainable by known to us algorithms for solving \rf{sa0}
the algorithm in \cite{agarwal2012stochastic,gaillard2017sparse} requires at least $s^2\ln[n]\over \kappa^2_\Sigma$  SMD iterations per stage, implying that the method in question can be used only
if the number of nonvanishing entries in the parameter vector is $O\left(\kappa_\Sigma\sqrt{N\over \ln n}\right)$\footnote{That being said, \cite{agarwal2012stochastic}, for instance, deals with {\em nonsmooth} stochastic optimization, so the scope of corresponding algorithms is much larger than the framework of smooth problems considered in this paper.} (recall that the corresponding limit is $O\left({N\kappa_\Sigma\over \ln[n]}\right)$ for Lasso \cite{raskutti2010restricted} and iterative thresholding procedures \cite{Barber1,Barber2}).

Our goal in the present paper is to provide a refined analysis of Stochastic Approximation algorithms for computing sparse solutions to \rf{prob0st} exploiting a variance reduction scheme utilizing in a special way smoothness of the problem objective.\footnote{In hindsight, the underlying idea can be seen as a generalization of the variance reduction device in \cite{bietti2017stochastic}.} It allows to build a new accelerated multistage Stochastic Approximation algorithm. To give a flavor of the results we present below, we summarize the properties of the proposed procedure---Stochastic Mirror Descent for Sparse Recovery (SMD-SR)---in the case of stochastic optimization problem \rf{sa0} associated with sparse linear regression estimation problem.
Let us assume that regressors $\phi_i$ are a.s. bounded, i.e., $\|\phi_i\|_\infty=O(1)$, the covariance matrix $\Sigma=\bE\{\phi_1\phi_1^T\}$ of regressors satisfies $\Sigma\succeq \kappa_\Sigma I$; we suppose that the noises $\sigma\xi_i$ are zero-mean with $\bE\{\xi_i^2\}\leq 1$, and that we are given $R<\infty$ and $x_0\in \bR^n$ such that $\bE\{\|x_0-x_*\|_1^2\}\leq R^2$.
\begin{itemize}\item The SMD-SR algorithm is organized in stages. On the $k$-th stage of the method we run $N_k$ iterations of the Stochastic Mirror Descent recursion and then ``sparsify'' the obtained approximate solution by zeroing out all but $s$ entries of largest amplitudes.
\item Stages of the algorithm are organized into two groups (phases). At the first (preliminary) phase we perform a fixed number $N_k=O\left({s\ln n\over \kappa_\Sigma } \right)$ of SMD iterations per stage to guarantee that the expected quadratic error $\bE\{\|\wh y_k-x_*\|_1^2\}$ of the sparse approximate solution $\wh y_k$ of the $k$-th stage is smaller than the expected error $\bE\{\|\wh y_{k-1}-x_*\|_1^2\}$ of the previous stage solution $y_{k-1}$ by a fixed factor. Thus, the error of the approximate solution after (total) $N$ iterations decreases linearly with the exponent proportional to ${\kappa_\Sigma\over s\ln n}$. When the expected quadratic error becomes $O\left({\sigma^2s^2\over \kappa_\Sigma}\right)$, we pass to the second (asymptotic) phase of the method.
\item During the stages of the asymptotic phase, the number of iterations per stage grows as $N_k=2^kN_0$ where $k$ is the stage index, and the expected quadratic error decreases as $O\left({\sigma^2s^2\ln n\over {\kappa_\Sigma^2}N}\right)$ where $N$ is total iteration count.
\end{itemize}
It may appear surprising that a stochastic algorithm converges linearly during the preliminary phase, when the component of the error due to the observation noise is small (for instance, it converges linearly in the  ``noiseless'' case, cf. \cite{nguyen2017linear}) eliminating fast the initial error; its rate of convergence is similar to that of the deterministic gradient descent algorithm, when ``full gradient observation'' $\nabla g(x)$ is available. On the other hand, in the asymptotic regime, the procedure attains the rate which is equivalent to the best known rates in this setting, and under the model assumptions which are close to the weakest known today \cite{Barber2,raskutti2010restricted}.
\par
The paper is organized as follows. The analysis of the SMD-SR in the general setting is  in Section~\ref{sec:1}. We define the general problem setting and introduce key notions used in the paper in Section~\ref{sec:probs}. Then in Section~\ref{sec:multis} we reveal the multistage algorithm and study its basic properties. Next, in Section \ref{sec:reliab} we show how sub-Gaussian confidence bounds for the error of approximate solutions can be obtained using an adopted analog of Median-of-Means approach. Finally, in Section \ref{sec:appl} we discuss the properties of the method and conditions in which it leads to ``small error'' solution when applied to sparse linear regression and low rank linear matrix recovery problems.


%
\section{Sparse solutions to stochastic optimization problem}\label{sec:1}
\subsection{Problem statement}\label{sec:probs}
	Let $E$ be a finite-dimensional real vector (Euclidean) space. Consider a Stochastic Optimization problem
	\be
	\min\limits_{x\in X} \left[\bE\{G(x,\omega)\}\right]
	\ee{prob1}
where  $X\subset E$ is a convex 
set with nonempty interior (a solid), $\omega$ is a random variable on a probability space $\Omega$ with distribution $P$, and $G:\;X\times \Omega\to \bR$. We suppose that the expected objective
$$g(x)=\bE\{G(x,\omega)\}$$ is finite for all $x\in X$ and is convex and differentiable on $X$.
Let $\|\cdot\|$ be a norm on $E$, and let $\| \cdot \|_*$ be the conjugate norm,
i.e.,
	\[
	\| s \|_* = \max\limits_x \{ \la s,x\ra: \; \|x\|\leq 1 \},
	\quad s \in E.
	\]
We suppose that gradient $\nabla g(\cdot)$ of  $g(\cdot)$
	is Lipschitz-continuous on $X$:
	\be
	\|\nabla g(x')-\nabla g(x)\|_*\leq \L\|x-x'\|,\qquad \forall \,x,x'\in X,
	\ee{Lipf}
that the problem is solvable with optimal value $g_*=\min_{x\in{X}}g(x)$. Furthermore, we suppose that the optimal solution $x_*$ to the problem is unique, and that $g(\cdot)$ satisfies {\em quadratic growth condition} on $X$ with respect to the Euclidean norm  $\|\cdot\|_2$ \cite{necoara2018linear}, i.e., for all $x\in X$
\be
g(x)-g_*\geq \half \lowkap \|x-x_*\|_2^2
\ee{quadgg}
where  $\|\cdot\|_2$ is the Euclidean norm: $\|z\|_2={\la z,z\ra}^{1/2}$.
	In what follows, we assume that we have at our disposal a {\em stochastic} (gray box) {\em oracle}---a device which can generate $\omega \sim P$ and compute, for any
	$x\in X$ a random unbiased estimation of $\nabla g(x)$.
From now on we make the following assumption about the structure of the gradient observation:
\paragraph{Assumption [S1].} {\em $G(\cdot,\omega)$ is 
differentiable on $X$ for almost all $\omega\in \Omega$, and\footnote{In what follows $\nabla G(\cdot,\omega)$ replaces notation $\nabla_x G(\cdot,\omega)$ for the gradient of $G$ w.r.t. the first argument. }
\[
\bE\{\nabla G(x,\omega)\}=\nabla g(x) \quad\mbox{and}\quad \bE\{\|\underbrace{\nabla G(x,\omega)-\nabla g(x)}_{=:\zeta(x,\omega)}\|^2_*\}\leq \varsigma^2(x),
	\quad \forall\,x\in X.
\]
Furthermore, there are $1\leq \varkappa,\varkappa'<\infty$ and $\L\leq \nu<\infty$ such that
the bound holds:}
\be
\varsigma^2(x)
\leq
\varkappa \nu 
[g(x)-g_*-\la\nabla g(x_*),x-x_*\ra]
+\varkappa'\underbrace{\bE\{ \|\zeta(x_*,\omega)\|_*^2\}}_{=:\varsigma_*^2}.
\ee{s2bound0}
\paragraph{Remarks.} Assumption {\bf S1} and, in particular, bound \rf{s2bound0} are essential to the subsequent developments and certainly merit some comments. We postpone the corresponding discussion to Section \ref{sec:appl} where we present several examples of observation models in which this assumption naturally holds. For now, let us consider a simple example of the Stochastic Optimization problem \rf{sa0} arising in sparse regression estimation where  regressors $\phi_i$ are a.s. bounded, i.e., $\|\phi_i\|_\infty\leq r<\infty$ with identity covariance matrix $\bE\{\phi_1\phi_1^T\}=I$, and noises $\sigma\xi_i$ are zero-mean with ``small'' variance.
In the situation in question, the error
$
\zeta(x,\omega)=\nabla G(x,\omega)-\nabla g(x)$, $\omega=[\phi,\xi]$, of the  stochastic oracle can be decomposed as in
\[
\zeta(x,\omega)=\underbrace{[\phi\phi^T-I](x-x_*)}_{=:\zeta_1(x,\omega)}+\underbrace{\sigma\xi\phi}_{=:\zeta_2(\omega)}.
\]
Note that the ``variance'' $\varsigma^2_1(x)$ of the first component 
satisfies
\[
\varsigma^2_1(x)=\bE\{\|\zeta_1(x,\omega)\|^2_\infty\}\leq 2(r^2+1) \|x-x_*\|_2^2\leq 4(r^2+1) (g(x)-g_*),
\]
 while the ``variance'' $\varsigma^2_2$ of the second,
\[\varsigma^2_2=\bE\{\|\zeta_2(\omega)\|^2_\infty\}=\sigma^2\bE\{\|\phi\|_\infty^2\}\leq \sigma^2 r^2,
\]
does not depend on $x$. As a result, the bound
\[
\varsigma^2(x)=\bE\{\|\zeta(x,\omega)\|^2_\infty\}\leq 4(r^2+1) \|x-x_*\|_2^2+2\sigma^2 r^2
\]
implies that in this case the stochastic gradient observation $\nabla G(x,\omega)$ satisfies Assumption {\bf S1} with $\varsigma_*^2=\sigma^2r^2$, $\nu=r^2+1$, $\kappa=8$ and $\kappa'=2$.

More generally, relation \rf{s2bound0} is rather characteristic to the case of smooth stochastic observation. Indeed, let us consider the situation where
the stochastic gradient $G(\cdot,\omega)$ itself is Lipschitz-continuous on $X$ with a.s. bounded Lipschitz constant $\L(\omega)$ with respect to the norm $\|\cdot\|$,
$
\L(\omega)\leq \nu$. In this case we have
\bse
\varsigma^2(x)&=&\bE\big\{\|\nabla G(x,\omega)-\nabla g(x)\|_*^2\big\}
\leq \Big(
\bE\big\{\|\nabla G(x,\omega)-\nabla G(x_*,\omega)\|_*^2\big\}^{1/2}\\
&&+\|\nabla g(x)-\nabla g(x_*)\|_*+
\bE\big\{\|\nabla G(x_*,\omega)-\nabla g(x_*)\|_*^2\big\}^{1/2}\Big)^2.
\ese
However, due to the Lipschitz continuity of $\nabla G(\cdot,\omega)$
\bse
G(x,\omega)- G(x_*,\omega)&\geq& \la\nabla G(x_*,\omega),x-x_*\ra+{(2\nu)^{-1}}\|\nabla G(x,\omega)-\nabla G(x_*,\omega)\|_*^2,
\ese
implying that
\bse
\varsigma^2(x)&\leq& \left([2\nu\bE\{G(x,\omega)-G(x_*,\omega)-\la\nabla G(x_*,\omega),x-x_*\ra\}]^{1/2}\right.\\&&+\left.
[2\nu(g(x)-g(x_*)-\la \nabla g(x_*),x-x_*\ra)]^{1/2}+\varsigma_*\right)^2\\&\leq& 16\nu[g(x)-g_*-\la \nabla g(x_*),x-x_*\ra]+2\varsigma_*^{2}.
\ese

\paragraph{Sparsity structure.}
In what follows we assume to be given a {\em sparsity structure} \cite{juditsky2014unified} on $E$---a family $\cP$ of projector mappings $P=P^2$ on $E$ with associated nonnegative weights $\pi(P)$. For a nonnegative real $s$ we set
\[
\cP_s=\{P\in \cP:\pi(P)\leq s\}.
\]
Given $s\geq 0$ we call $x\in E$ {\em $s$-sparse} if there exists $P\in \cP_s$ such that $Px=x$. We will make the following standing assumption.
\paragraph{Assumption [S2]} {\em The optimal solution $x_*$ to problem \rf{prob1} is $s$-sparse.

Furthermore, given $x\in X$ one can efficiently compute a ``sparse approximation'' of $x$---an optimal solution $x_s=\sparse(x)$ to the optimization problem
\be
\min \|x-z\|_2\;\;\mbox{over $s$-sparse $z\in X$}.
\ee{sparseapp}
Moreover,  for any $s$-sparse $z\in E$ the norm $\|\cdot\|$ satisfies $\|z\|\leq \sqrt{s}\|z\|_2$. }
\par
In what follows we refer to $x_s$ as ``sparsification of $x$.'' We are mainly interested in the following ``standard examples'':
\begin{enumerate}
\item ``Vanilla'' sparsity: in this case $E=\bR^n$ with the standard inner product, $\cP$ is comprised of projectors on all coordinate subspaces of $\bR^n$, $\pi(P)=\rank(P)$, and $\|\cdot\|=\|\cdot\|_1$.
    \par Assumption {\bf S2} clearly holds, for instance, when $X$ is orthosymmetric, e.g., a ball of $\ell_p$-norm on $\bR^n$, $1\leq p\leq \infty$.
\item Group sparsity: $E=\bR^n$, and we partition the set $\{1,...,n\}$ of indices into $K$ nonoverlapping subsets $I_1,...,I_K$, so that to every $x\in \bR^n$ we associate blocks $x^k$ with corresponding indices in $I_k,\,k=1,...,K$. Now $\cP$ is comprised of projectors $P=P_I$ onto subspaces $E_I=\{[x^1,...,x^K]\in \bR^n:\,x^k=0\, \forall k\notin I\}$ associated with subsets $I$ of the index set $\{1,...,K\}$. We set $\pi(P_I)=\card I$, and define $\|x\|=\sum_{k=1}^K\|x_k\|_2$---{\em block $\ell_1/\ell_2$-norm.}
    \par Same as above, Assumption {\bf S2} holds in this case when $X$ is ``block-symmetric,''  for instance, is a ball of block norm $\|\cdot\|$.
\item Low rank sparsity structure: in this example $E=\bR^{p\times q}$ with, for the sake of definiteness, $p\geq q$, and the Frobenius inner product. Here $\cP$ is the set of mappings $P(x)=P_\ell x P_r$ where  $P_\ell$ and $P_r$ are, respectively, $q\times q$ and $p\times p$ orthoprojectors, and $\|\cdot\|$ is the nuclear norm $\|x\|=\sum_{i=1}^q\sigma_i(x)$ where $\sigma_1(x)\geq \sigma_2(x)\geq ...\geq \sigma_q(x)$ are singular values of $x$.
    \par
    In this case Assumption {\bf S2} holds due to the Eckart–Young approximation theorem, it suffices that $X$ is a ball of a Schatten norm
    $\|x\|_r=\left(\sum_{i=1}^q \sigma^r_i(x)\right)^{1/r}$, $1\leq r\leq \infty$.
\end{enumerate}
Our objective is to build approximate solutions $\wh{x}_N$ to problem \rf{prob1} utilizing $N$ queries to the stochastic oracle. We quantify the performance of such solutions on the class $
\X=\X(X,\L,\,...,\,\P,s)$ of Sparse Stochastic Optimization problems \rf{prob1} described in the beginning of this section satisfying Assumptions {\bf S1} and {\bf S2}, with domain $X$,  by the following worst-case over $\X$ risk measures:
\begin{itemize}
\item{\em Recovery risks:} maximal over $\X$ expected squared error
\[
\risk_{|\cdot|}(\wh x|\X)=\sup_{\X}\bE\{|\wh x-x_*|^2\}^{1/2}
\]
where $|\cdot|$ stands for $\|\cdot\|_2$- or $\|\cdot\|$-norm,
and {\em $\epsilon$-risk} of recovery---the smallest maximal over $\X$ radius of $(1-\epsilon)$-confidence ball of norm $|\cdot|$ centered at $\wh x$:
\[
\risk_{|\cdot|,\epsilon}(\wh x|\X)=\inf \left\{r:\sup_{\X}\Prob\{|\wh x-x_*|\geq r\}\leq \epsilon\right\}
\]
\item{\em Prediction risks:} maximal over $\X$ expected suboptimality
\[
\risk_{g}(\wh x|\X)=\sup_{\X}\bE\{g(\wh x)\}-g_*,
\]
of $\wh x$ and the smallest maximal over $\X$ $(1-\epsilon)$-confidence interval 
\be
\risk_{g,\epsilon}(\wh x|\X)=\inf \left\{r:\sup_{\X}\Prob\{g(\wh x)-g_*\geq r\}\leq \epsilon\right\}.
\ee{geps}
\end{itemize}
In what follows, we use a generic notation $c$ and $C$ for absolute constants; notation $a \lesssim b$ means that the ratio $a/b$ is bounded by an absolute constant.

\subsection{Stochastic Mirror Descent algorithm}\label{sec:pstat}
\paragraph{Notation and definitions.}
Let  $\vartheta:\,E\to\bR$ be a continuously differentiable  convex function which is strongly convex with respect to the norm $\|\cdot\|$, i.e.,
\[
\la\nabla \vartheta(x)-\nabla \vartheta(x'),x-x'\rangle\ge \|x-x'\|^2,\quad \forall x,x'\in E.
\]
From now on, w.l.o.g. we assume that $\vartheta(x)\geq \vartheta(0)=0$. We say that $\Theta$ is the constant of quadratic growth of $\vartheta(\cdot)$ if
\[\forall x\in E\;\vartheta(x)\leq \Theta\|x\|^2.
\]
Clearly, $\Theta\geq \half$. If, in addition, $\Theta$ is ``not too large,'' and for any $x\in X$, $a\in E$ and $\beta>0$ a high accuracy solution to the minimization problem
	\[
	\min_{z\in X} \{\langle a,z\rangle+\beta \vartheta(z-x) \}
	\]
	can be easily computed, following \cite{juditsky2011optimization,juditsky2014deterministic,nemirovski2009robust,nesterov2013first} we say that {\em distance-generating function (d.-g.f.) $\vartheta$} is ``prox-friendly.'' We present choices of prox-friendly d.-g.f.'s relative to the norm used in application sections. \par
 We also utilize associated Bregman divergence
	\[
	V_{x_0}(x,z)=\vartheta(z-x_0)-\vartheta(x-x_0)-\la\nabla \vartheta(x-x_0),z-x\ra,
	\quad\forall\, z, x, x_0\in{X}.
	\]
For $Q\in \bR^{p\times q}$ we denote
\[\|Q\|_\infty=\max_{ij}|[Q]_{ij}|;
\] for symmetric positive-definite $Q\in \bR^{n\times n}$ and $x\in \bR^n$ we denote
\[\|x\|_Q=\sqrt{x^TQx}.\]
\paragraph{Stochastic Mirror Descent algorithm.}
For $x,x_0\in X$, $u\in E$, and $\beta>0$ consider the {\em proximal mapping}
	\be
	\Prox_{\beta}(u,x;x_0)&:=&\argmin_{z\in X}\big\{\langle  u,z\rangle +\beta V_{x_0}(x,z)\big\}\nn
 &=&\argmin_{z\in X}\big\{\langle  u-\beta \la\nabla \vartheta(x-x_0),z\rangle+\beta\vartheta(z-x_0)\big\}.
	\ee{prox}
	For $i=1,2,\dots$,
	consider {\em Stochastic Mirror Descent} recursion, cf.  \cite{juditsky2011optimization,nemirovski2009robust,lan2012optimal},
	\be	x_{i}&=&\Prox_{\beta_{i-1}}(\nabla G(x_{i-1},\omega_i),x_{i-1};x_0),\;\;\;x_0\in X,
	\ee{eq:md1}
Here $\beta_i>0$, $i=0,1,\dots$, is a stepsize parameter to be defined later,
	and $\omega_1,\omega_2,\dots$ are independent identically distributed (i.i.d.) realizations of random variable $\omega$,
	corresponding to the oracle queries at each step of the algorithm.
	
The approximate solution to problem \rf{prob1} after $N$ iterations is defined as weighted average
\be
\widehat{x}_N=\left[\sum_{i=1}^{N}\beta^{-1}_{i-1}\right]^{-1}\sum_{i=1}^{N} \beta_{i-1}^{-1}x_i.
\ee{eq:asol}	
	The next result describes some useful properties of the recursion \rf{eq:md1}.
\begin{proposition}\label{pr:myprop2}
Suppose that SMD algorithm is applied to problem \rf{prob1} in the situation described in this section. We assume that Assumption {\bf S1} holds and that initial condition $x_0\in X$ is independent of $\omega_i$, $i=1,2,...$ and such that $\bE\{\|x_0-x_*\|^2\}\leq R^2$; we use constant stepsizes \[
\beta_i\equiv\beta\geq2\varkappa\nu 
, \;\;i=1,2,..., m.
\] Then approximate solution
$\wh x_m={1\over m}\sum_{i=1}^m x_{i}$ after $m$ steps of the algorithm satisfies
\be
\bE\{g(\wh x_{m})\}-g_*\leq
{2R^2 \over m}\left({\Theta\beta}+{\varkappa\nu^2\over 2\beta}\right)+
{2\varkappa'\varsigma_*^2\over \beta}.
\ee{bouprop1}
\end{proposition}
\subsection{Multistage SMD algorithm}\label{sec:multis}
We assume to be given $R<\infty$ and $x_0\in X$ such that $\|x_*-x_0\|\leq R$, along with problem parameters $\varkappa,\varkappa',\nu,\varsigma^2_*,\lowkap$ and an upper bound $\bar s$ for signal sparsity.
We are using the Stochastic Mirror Descent algorithm and  apply the multistage modification of \cite{juditsky2014deterministic,OML2011} to improve its accuracy bounds.
The proposed Stochastic Mirror Descent algorithm for Sparse Recovery (SMD-SR) works in stages---runs of the Stochastic Mirror Descent algorithm followed by subsequent ``sparsification'' of the approximate solution delivered by the SMD. The stages are split into two groups---phases---corresponding to two different regimes of the method. This organization of the algorithm allows to treat differently two components in the bound \rf{bouprop1} for the error of the Stochastic Mirror Descent algorithm.

During the first {\em preliminary} phase of the algorithm, the first term in the right-hand side of \rf{bouprop1} is dominant. This term is proportional to the bound $R^2$ on the expected squared $\ell_1$-norm of the error of the initial solution, and decreases as $1/m$ where $m$ is the iteration count.
During the stages of the preliminary phase, the stepsize parameter $\beta$ and the number of iterations per stage are set constant in such a way that the bound for the expected squared
error of the approximate solution decreases  by a constant factor at the end of the stage. Therefore, during this phase, the error of approximate solution converges linearly as a function of the total number of calls to stochastic oracle.

Preliminary phase terminates when the first term in the error bound \rf{bouprop1} becomes dominated with the second, independent of the initial error of the algorithm. During the second {\em asymptotic} phase of the method, the choice of the stepsize parameter and the length of the stage are ``standard'' for multistage Stochastic Mirror Descent (cf., e.g., \cite{juditsky2014deterministic}) and the method converges sublinearly, with the ``standard'' rate $O(1/N)$ where $N$ is the total number of oracle calls.

\paragraph{Algorithm 1 [SMD-SR]}
\begin{enumerate}\item
\underline{Preliminary phase}
\begin{description}\item[]{\em Initialization:} Set $y_0=x_0\in X$, $R_0=R$,
\be
\beta_0=2\varkappa\nu, \;\;m_0=\left\lceil 16\lowkap^{-1}\bar s (8\Theta\varkappa+1)\nu\right\rceil
\ee{betapr}
(here $\lceil a\rceil$ stands for the smallest integer greater or equal to $a$).
Put
 \[
\overline K= \left\lceil \ln _2\left({ R_0^2\lowkap\nu\varkappa \over 32\varsigma^2_*\bar s\varkappa'}\right)
\right\rceil
 \]
 and run
\[K=\min\left\{\left\lfloor {N\over m_0}\right\rfloor,\overline K\right\}
\] stages of the preliminary phase (here $\lfloor a\rfloor$ stands for the ``usual'' integer part -- the largest integer less or equal to $a$).
\item[]{\em Stage $k=1,..., K$:} Compute approximate solution $\wh x_{m_0}(y_{k-1},\beta_0)$ after $m_0$ iterations of SMD algorithm with constant stepsize parameter $\beta_0$, corresponding to the initial condition $x_0=y_{k-1}$. 
Then define $y_k$ as ``$s$-sparsification'' of $\wh x_{m_0}(y_{k-1},\beta_0)$, i.e.,
$y_k=\sparse(\wh x_{m_0}(y_{k-1},\beta_0))$.
\item[]{\em Output:} define $\wh{y}^{(1)}=y_K$ and $\wh{x}^{(1)}=\wh x_{m_0}(y_{K-1},\beta)$ as approximate solutions at the end of the phase.
\end{description}
\item
Set $M=N-m_0\overline K$ and
\[m_k=\left\lceil 512{\bar s\Theta\nu\varkappa\over \lowkap}2^k\right\rceil,\;\;k=1,...
\]
If $m_1>M$ terminate and output $\wh{y}_N=\wh{y}^{(1)}$ and $\wh{x}_N=\wh{x}^{(1)}$ as approximate solutions by the procedure; otherwise, continue with stages of the asymptotic phase.
\par
\underline{Asymptotic phase}
\begin{description}\item[]{\em Initialization:} Set
\[K'=\max\left\{k:\,\sum_{i=1}^km_i\leq M\right\},
\]
$y'_0=\wh{y}^{(1)}$, and $\beta_k=2^k\nu\varkappa$, $k=1,...,K'$. 
\item[]{\em Stage $k=1,..., K'$:} Compute $\wh x_{m_k}(y'_{k-1},\beta_k)$; same as above, define $y'_k=\sparse(\wh x_{m_k}(y'_{k-1},\beta_k))$.
\item[]{\em Output:} After $K'$
stages, output $\wh{y}_N=y'_{K'}$ and $\wh{x}_N=\wh x_{m_{K'}}(y'_{K'-1},\beta_{K'})$.
\end{description}

\end{enumerate}
Properties of the proposed procedure are summarized in the following statement.
\begin{theorem}\label{cor:mycor01}
In the situation of this section, suppose that $N\geq m_0$ so at least one preliminary stage of Algorithm 1 is completed. Then 
approximate solutions $\wh x_N$ and $\wh y_N$ produced by the algorithm satisfy
 \be\label{1finb}
\risk_{g}(\wh x_N|\X)
 &\leq&{\lowkap R^2\over \bar s}\exp\left\{-{c N\lowkap\over \Theta\varkappa\bar s\nu}\right\}+
C{\varsigma^2_*\bar s\varkappa'\Theta\over \lowkap N},\\
\risk_{\|\cdot\|}(\wh y_N|\X)&\leq &\sqrt{2s} \risk_{\|\cdot\|_2}(\wh y_N|\X)
 \leq\sqrt{ 8s}\risk_{\|\cdot\|_2}(\wh x_N|\X)
 \nn&\lesssim &
 R\exp\left\{-{c N\lowkap\over \Theta\varkappa\bar s\nu}\right\}+{\varsigma_*\bar s\over \lowkap} \sqrt{\Theta \varkappa'\over N}.
 \ee{1fina}
\end{theorem}
\subsection{Enhancing the reliability of SMD-SR solutions}\label{sec:reliab}
In this section, our objective is to build approximate solutions to problem \rf{prob1} utilizing Algorithm 1 which obey ``sub-Gaussian type'' bounds on their $\epsilon$-risks.
Note that bounds \rf{1finb} and \rf{1fina} of Theorem \ref{cor:mycor01} do allow only for Chebyshev-type bounds for risks of $\wh y_N$ and $\wh x_N$. Nevertheless, their confidence can be easily improved by applying, for instance, an adapted version of ``median-of-means'' estimate \cite{nemirovskii1979complexity,minsker2015geometric}.
\paragraph{Reliable recovery utilizing geometric median of SMD-SR solutions.} Suppose that available sample of length $N$ can be split into $L$ independent samples of length $M=N/L$ (for the sake of simplicity let us assume that $N$ is a multiple of $L$). We run Algorithm 1 on each subsample thus obtaining $L$ independent recoveries $\wh x^{(1)}_M,...,\wh x^{(L)}_M$ and compute ``enhanced solutions'' using an  aggregation procedure of geometric median-type. Note that we are in the situation where Theorem \ref{cor:mycor01} applies, meaning that approximate solutions $\wh x^{(1)}_M,...,\wh x^{(L)}_M$  satisfy
\be
 \forall \ell\;\;\;\bE\{g(\wh{x}^{(\ell)}_M)\}-g_*\leq\tau_M^2:={\lowkap R^2\over \bar s}\exp\left\{-{cM\lowkap\over \Theta\varkappa\bar s\nu}\right\}+C
{\varsigma^2_*\bar s\varkappa'\Theta\over \lowkap M},
\ee{Mbound1}
and so
\be
 \forall \ell\;\;\;\bE\{\|\wh x^{(\ell)}_M-x_*\|^2_2\}\leq \theta_M^2:={2\over \lowkap}\tau_M^2
\lesssim{R^2\over \bar  s}\exp\left\{-{c M\lowkap\over \Theta\varkappa \bar s\nu}\right\}+
{\Theta \varkappa'\varsigma_*^2\bar s\over \lowkap^2M}.
\ee{Mbound2}
We are to select among $\wh x^{(\ell)}_M$ the solution which attains similar bounds ``reliably.''

\begin{enumerate}
\item The first reliable solution $\wh x_{N,1-\epsilon}$ of $x_*$ is a ``pure'' geometric median of $\wh x^{(1)}_M,...,\wh x^{(L)}_M$: we put
\be
\wh x_{N,1-\epsilon}\in \Argmin_{x} \sum_{\ell=1}^L \|x-\wh x^{(\ell)}_M\|_2,
\ee{l2med}
and then define $\wh y_{N,1-\epsilon}=\sparse(\wh x_{N,1-\epsilon})$. \footnote{Reliable solution we consider here explicitly depend on the confidence level; for instance,  parameter $L$ in the definition \rf{l2med} of $\wh x_{N,1-\epsilon}$ will be chosen depending on $\epsilon$. Hence, the presence of the index  $1-\epsilon$ in the notation of these estimates.}
\par
Computing reliable solutions $\wh x_{N,1-\epsilon}$ and $\wh y_{N,1-\epsilon}$ as optimal solutions to \rf{l2med} amounts to solving a nontrivial optimization problem. A simpler reliable estimation can be computed by replacing the geometric median $\wh x_{N,1-\epsilon}$ by its ``empirical counterparts'' (note that, number $L$ of solutions to be aggregated is not large---it is typically order of $\ln[1/\epsilon]$).
\item
We can replace $\wh x_{N,1-\epsilon}$ with
\[
\wh x'_{N,1-\epsilon}\in \Argmin_{x\in \{\wh x^{(1)}_M,...,\wh x^{(L)}_M\}} \sum_{\ell=1}^L \|x-\wh x^{(\ell)}_M\|_2
\]
and compute its sparse approximation $\wh y'_{N,1-\epsilon}=\sparse(\wh x'_{N,1-\epsilon})$.
\item Another reliable solution (with slightly better guarantees) was proposed in \cite{hsu2014heavy}.
Let $i\in\{1,...,L\}$, we set \[r_{ij}=\|\wh x^{(i)}_M-\wh x^{(j)}_M\|_2\] and denote $r^i_{(1)}\leq r^i_{(2)}\leq ... \leq r^i_{(L-1)}$ corresponding order statistics (i.e., $r_{i\cdot}$'s sorted in the increasing order). We define reliable solution $\wh x''_{N,1-\epsilon}=\wh x^{(\wh i)}_M$ where
\be
\wh i\in \Argmin_{i\in \{1,...,L\}} r^i_{\rceil L/2\lceil}
\ee{3rdmed}
(here  $\rceil a\lceil=\lfloor a\rfloor+1$ stands for the smallest integer strictly greater than $a$),
and put $\wh y''_{N,1-\epsilon}=\sparse(\wh x''_{N,1-\epsilon})$.
\end{enumerate}
\begin{theorem}\label{cor:reli}
Let $\epsilon\in (0,\four]$, and let $\overline{x}_N$ (resp. $\overline{y}_N$) be one of reliable solutions $\wh x_{N,1-\epsilon},\wh x'_{N,1-\epsilon}$ and $\wh x''_{N,1-\epsilon}$ (resp., $\wh y_{N,1-\epsilon},\wh y'_{N,1-\epsilon}$ and $\wh y''_{N, 1-\epsilon}$) described above using
$L=\lceil\alpha\ln [1/\epsilon]\rceil$\footnote{The exact value of the numeric constant $\alpha$ is specific for each construction,  and can be retrieved from the proof of the theorem.} independent
 approximate solutions $\wh x^{(1)}_M,...,\wh x^{(L)}_M$ by Algorithm 1.
When $N\geq Lm_0$ we have
\be
\risk_{\|\cdot\|,\epsilon}(\overline y_N|\X)&\leq& \sqrt{2s}\risk_{\|\cdot\|_2,\epsilon}(\overline y_N|\X)\leq 2\sqrt{2s}\risk_{\|\cdot\|_2,\epsilon}(\overline x_N|\X)
\nn&\lesssim&
R\exp\left\{-{c N\lowkap\over \Theta\varkappa\bar s\nu\ln[1/\epsilon]}\right\}+{\varsigma_*\bar s\over \lowkap}\sqrt{\Theta \varkappa'\ln[1/\epsilon]\over N}.
\ee{yhaterisk}
\end{theorem}
\paragraph{Remark.} Notice that the term $\ln[1/\epsilon]$ enters the bound \rf{yhaterisk} as a multiplier which is typical for accuracy estimates of solutions which relies upon median to enhance confidence; at the moment, we do not know if this dependence on reliability tolerance parameter may be improved.
\paragraph{Reliable solution aggregation.}
Let us assume that two independent observation samples of lengths $N$ and $K$ are available. In the present approach, we use the first sample to compute, same as in the construction presented above, $L$ independent approximate SMD-SR solutions $\wh x^{(\ell)}_M,\,\ell=1,...,L$, $M=N/L$. Then we ``aggregate'' $\wh x^{(1)}_M,...,\wh x^{(L)}_M$---select the best of them in terms of the objective value $g(\wh x^{(\ell)}_M)$ by computing reliable estimations of differences $g(\wh x^{(i)}_M)-g(\wh x^{(j)}_M)$ using observations of the second subsample.
\par
The proposed procedure for reliable selection of the ``best'' solution $\wh x^{(\ell)}_M$ is as follows.
\paragraph{Algorithm 2 [Reliable aggregation]}
\begin{itemize}\item[]{\em Initialization:} Algorithm parameters are $\epsilon\in(0,\half]$,
$L'\in \bZ_+$ and $m=K/L'$
(for the sake of simplicity we assume, as usual, that $K=mL'$). We assume to be given $L$ points $\wh x^{(1)}_M,...,\wh x^{(L)}_M$ (approximate solution of the first step).
\\We compute $\wh x''_{N,1-\epsilon}=\wh x^{(\wh i)}_{M}$ the reliable solution as defined in \rf{3rdmed} and denote $\wh I=\{i_1,...,i_{\rceil L/2\lceil}\}$, the set of indices of $\rceil L/2\lceil$ closest to $\wh x''_{N,1-\epsilon}$ in the Euclidean norm points among $\wh x^{(1)}_{M},...,\wh x^{(L)}_M$.
\item[]{\em Comparison procedure:} We split the (second) sample $\omega^K$ into $L'$ independent subsamples $\omega^\ell$, $\ell=1,...,L'$ of size $m$.
For all $i\in \wh I$  we compute the index \[
\wh v_{i}=\max_{j\in \wh I,\,j\neq i} \left\{\med_{\ell}[\wh v^\ell_{ji}]-\rho_{ij}\right\}
\] where
 \[
 \wh v^\ell_{ji}={1\over m}\sum_{k=1}^m\big\langle \nabla G(\wh x^{(j)}_M+t_k(\wh x^{(i)}_M-\wh x^{(j)}_M),\omega^{\ell}_k),\wh x^{(i)}_M-\wh x^{(j)}_M\big\rangle,\;\;\ell=1,...,L',
 \]
are estimates of $v_{ji}=g(\wh x^{(i)}_M)-g(\wh x^{(j)}_M)$, $t_k={2k-1\over 2m}$, $k=1,...,m$, 
 and coefficients $ \rho_{ij}>0$ to be defined depend on $r_{ij}=\|\wh x^{(i)}_M-\wh x^{(j)}_M\|_2$.
\item{Output:} We say that {\em $x^{(i)}_M$ is admissible} if $\wh v_i\leq 0$. When the set of admissible $\wh x^{(i)}_M$'s is nonempty we define the procedure output $\overline x_{N+K,1-\epsilon}$ as one of admissible  $\wh x^{(i)}_M$'s, and  define $\overline x_{N+K,1-\epsilon}=\wh x^{(1)}_M$  otherwise.
\end{itemize}
Now, consider the following (cf. Assumption {\bf S1})
\paragraph{Assumption [S3].}
{\em There are $1\leq \chi,\chi'<\infty$ such that for any $x\in X$ and $z\in E$
the following bound holds:
\be
\bE\{\la \zeta(x,\omega),z\ra^2\}\leq
\|z\|^2_2[\chi\L_2  (g(x)-g_*)+\chi'\varsigma_*^2]
\ee{s3bound0}
where $\L_2$ is the Lipschitz constant of the gradient $\nabla g$ of $g$ with respect to the Euclidean norm,
\[
\|\nabla g(x')-\nabla g({x''})\|_2\leq \L_2\|x'-x''\|_2,\;\;\forall x',x''\in X.
\]}
Let now $\overline \X$ be the class of Sparse Stochastic Optimization problems as described in Section \ref{sec:probs} satisfying Assumptions {\bf S1}--{\bf S3}, with domain $X$. Assume that
risk $\risk_{g,\epsilon}(\cdot|\overline\X )$ is defined as in \rf{geps} with $\X$ replaced with $\overline \X$.
\begin{theorem}\label{cor:reli2} Let Assumption {\bf S3} hold, and let $\tau_M$ and $\theta_M$ be as in \rf{Mbound1} and \rf{Mbound2} respectively. Further, in the situation of this section, let $\epsilon\in (0,\half]$, $L=\lceil \alpha\ln [1/\epsilon]\rceil$ for large enough $\alpha$, and let $\overline x_{N+K,1-\epsilon}$ be an approximate solution by Algorithm 2
 in which we set $L'\geq \Big\lceil 7\ln [2/\varepsilon]\Big\rceil$ and
\[
\rho_{ij}=2r_{ij}\sqrt{\L_2\chi\over m}(\gamma(r_{ij})+\tau_M)+2r_{ij}\varsigma_*\sqrt{\chi'\over m}
\]
where
\be
\gamma(r)=\left(\Big[4r\sqrt{\chi\L_2\over m}+\tau_M\Big]^2+4r\zeta_*\sqrt{\chi'\over m}\right)^{1/2}.
\ee{***gam}
Then
\[
\risk_{g,\epsilon}(\overline x_{N+K,1-\epsilon}|\overline\X)\leq \bar\gamma^2:=\gamma^2(8\theta_M),
\]
In particular, when $K=mL'\geq
c\max \left\{{\chi \L_2\ln[1/\epsilon]\over \lowkap} ,
{N\chi'\over \Theta\varkappa'\bar s}\right\}
$ for an appropriate absolute $c>0$,  one has
\[
\risk_{g,\epsilon}(\overline x_{N+K,1-\epsilon}|\overline\X)\lesssim
{\lowkap R^2\over \bar s}\exp\left\{-{c N\lowkap\over \Theta\varkappa\bar s\nu\ln[1/\epsilon]}\right\}+{\varsigma_*^2\bar s\Theta \varkappa'\ln[1/\epsilon]\over \lowkap N}.
\]
\end{theorem}
\section{Applications}\label{sec:appl}
%
\subsection{Sparse generalized linear regression by stochastic approximation}\label{sec:sr}
Let us consider the problem of recovery of a sparse signal $x_*\in \bR^n$, $n\geq 3$, from independent and identically distributed observations
\be
\eta_i=\myu(\phi_i^Tx_*)+\sigma\xi_i,\;\;\;i=1,2,...,N,
\ee{sparselin}
where ``activation'' $u:\bR\to \bR$, $\phi_i\in \bR^n$ and $\xi_i\in \bR$ are mutually independent and such that
$\bE\{\phi_i\phi_i^T\}=\Sigma$, $\kappa_\Sigma I\preceq\Sigma$, and $\|\Sigma\|_\infty\leq \upsilon$, with known $\kappa_\Sigma>0$ and $\upsilon$;\footnote{Recall that for a matrix $Q$ we denote $\|Q\|_\infty=\max_{ij}|[Q]_{ij}|$.} we also assume that 
$\bE\{\xi_i\}=0$ and $\bE\{\xi_i^2\}\leq 1$.
\par
We suppose that $x_*$ is {\em $s$-sparse} and that we are given a convex and closed subset $X$ of $\bR^n$ (e.g., a large enough ball of $\ell_1$- or $\ell_2$-norm centered at the origin) such that $x_*\in X$, along with $R<\infty$ and $x_0\in X$ such that $\|x_*-x_0\|_1\leq R$. Furthermore, the mapping $\myu(\cdot)$ is assumed to be known, strongly monotone and Lipschitz continuous, i.e., for some $0< \ul\leq \ol$ and  all $t\geq t'$
\be
\ul(t-t')\leq \myu(t)-\myu(t')\leq \ol(t-t').
\ee{ucond}
\par We are about to apply Stochastic Optimization approach described in Section \ref{sec:1}. To this end, let $\myU$ be the primitive of $u$, i.e., $\myU'(t)=\myu(t)$, and let us consider
the Stochastic Optimization problem
\be
\min_{x\in X}\left\{g(x)=\half\bE\big\{\underbrace{\myU(\phi^Tx)-\phi^Tx \eta }_{=:G(x,\omega=[\phi,\eta])}\big\}\right\}.
\ee{SOP}
Note that $x_*$ is the unique optimal solution to the above problem. Indeed, observe that
$\nabla G(x,\omega)=\phi (\myu(\phi^Tx)-\eta)$ and $\bE_\xi\{\eta\}=\myu(\phi^Tx_*)$. We have $\nabla g(x_*)=0$; furthermore,
\bse
g(x)-g(x_*)&=&\int_0^1 \nabla g(x_*+t(x-x_*))^T(x-x_*)dt\\&=&
\int_0^1 \bE\Big\{\phi [\myu(\phi^T(x_*+t(x-x_*))-\myu(\phi^Tx_*)]\big\}^T(x-x_*)dt\\
\mbox{[by \rf{ucond}]}&
\geq &
\int_0^1 \ul \bE\big\{[\phi^T(x-x_*)]^2\big\}tdt=
\half \ul \|x-x_*\|^2_\Sigma\geq \half \ul\kappa_\Sigma \|x-x_*\|^2_2,
\ese
and we conclude that $g$ is quadratically minorated with parameter $\lowkap=\ul \kappa_\Sigma$.

We set $\|\cdot\|=\|\cdot\|_1$ with $ \|\cdot\|_*=\|\cdot\|_\infty$, and
%
we use ``$\ell_1$-proximal setup'' of the SMD-SR algorithm with  quadratically growing  for $n>2$ distance-generating function (cf. \cite[Theorem 2.1]{nesterov2013first})
\[
\vartheta(x)=\half e\ln(n)\, n^{(p-1)(2-p)/p }\|x\|_p^2,\;\;p=1+{1\over \ln n},
\]
the corresponding $\Theta$ satisfying $\Theta\leq \half e^2\ln n$.
\par
Note that, due to \rf{ucond}, for all $z\in \bR^n$ such that $\|z\|_1\leq 1$
\bse
|z^T(\nabla g(x)-\nabla g(x'))|&=&\big|\bE\big\{\phi^Tz(\myu(\phi^Tx)-\myu(\phi^Tx'))\big\}\big|\leq \ol \bE\{|\phi^Tz|\,|\phi^T(x-x')|\}\\
&\leq&
\ol \bE\{(\phi^Tz)^2\}^{1/2}\bE\big\{(\phi^T(x-x'))^2\big\}^{1/2}\leq \ol\upsilon^{1/2} \|x-x'\|_\Sigma,
\ese
i.e., $\|\nabla g(x)-\nabla g(x')\|_\infty\leq\ol\upsilon^{1/2} \|x-x'\|_\Sigma$. Thus,
\bse
\varsigma(x)&=&\bE\big\{\|\nabla G(x,\omega)-\nabla g(x)\|_\infty^2\big\}^{1/2}
\leq \bE\big\{[\|\phi(\myu(\phi^Tx)-\myu(\phi^Tx_*))-\nabla g(x)\|_\infty+\|\phi\xi\|_\infty]^2\big\}^{1/2}\\
&\leq &\ol\bE\big\{\|\phi\|^2_\infty(\phi^T(x-x_*))^2\big\}^{1/2}+ \ol\upsilon^{1/2} \|x-x'\|_\Sigma+\mynu\sigma
\ese
where $\mynu=\bE\{\|\phi\|_\infty^2\}^{1/2}$.
In other words, Assumption {\bf S1} holds whenever
\be
\varsigma^2(x)\leq \left(\ol\bE\big\{\|\phi\|^2_\infty(\phi^T(x-x_*))^2\big\}^{1/2}+ \ol\upsilon^{1/2} \|x-x'\|_\Sigma+\mynu\sigma\right)^2\leq \varkappa\nu(g(x)-g_*)+\varkappa'\varsigma_*^2
\ee{s2bound00}
which is the case if, for instance,
\be
\ol^2\bE\big\{\|\phi\|^2_\infty(\phi^T(x-x_*))^2\big\}\lesssim\nu\ul\|x-x'\|^2_\Sigma.
\ee{s2bound000}
and $\varsigma_*$ satisfies $\varsigma_*^2\geq \mynu^2\sigma^2$.
\paragraph{Remark.}
In the special case of $\myu(t)=t$, one has
\bse
g(x)&=&\bE\big\{\underbrace{\half(\phi^Tx)^2-\phi^Tx\eta}_{=G(x,\omega)}\big\}=\half \bE\big\{[\phi^T(x_*-x)]^2-(\phi^Tx_*)^2\big\}\\&=&\half (x-x_*)^T\Sigma(x-x_*)-\half x_*^T\Sigma x_*
=\half \|x-x_*\|_\Sigma^2-\half \|x_*\|^2_\Sigma
\ese
with $
\nabla g(x)=\Sigma (x-x_*)=\bE\big\{\underbrace{\phi\phi^T(x-x_*)-\sigma\xi\phi}_{=:\nabla G(x,\omega)}\big\}.$
In this case,
\[
\zeta(x,\omega)=\nabla G(x,\omega)-\nabla g(x)=[\phi\phi^T-\Sigma](x-x_*)-\sigma\xi\phi,
\]
and
\[
\varsigma^2(x)=
\bE\big\{\|[\phi\phi^T-\Sigma] (x-x_*)-\sigma\xi\phi\|_\infty^2\big\}.
\]
In this situation, Assumption {\bf S1} simplifies to
\[ \bE\big\{\|[\phi\phi^T-\Sigma] (x-x_*)-\sigma\xi\phi\|_\infty^2\big\}\leq
\half \varkappa \nu \|x-x_*\|_\Sigma^2+\varkappa'\varsigma_*^2
\]
which is satisfied with $\varsigma_*^2=\mynu^2\sigma^2$ whenever $\bE\big\{\|\phi\|^2_\infty(\phi^T(x-x_*))^2\big\}\lesssim\nu\|x-x'\|^2_\Sigma$.

Our present goal is to  describe the properties of approximate solutions by Algorithm 1 when applied to the optimization problem in \rf{SOP}.  We assume that the problem parameters---values $\varkappa,\,\nu,\,\kappa_\Sigma, \,\sigma^2$ and an upper bound $\bar s$ on sparsity of $x_*$---are known. We consider the following performance characteristics of approximate solutions $\wh x$---analogues of risks measures defined in Section \ref{sec:probs}---in our present situation:
\begin{itemize}
\item{\em Recovery risks:} maximal over $x_*\in X$ expected squared error
\be
\risk_{|\cdot|}(\wh x|X)=\sup_{x_*\in X}\bE\{|\wh x-x_*|^2\}^{1/2}
\ee{risk1}
where $|\cdot|$ stands for $\|\cdot\|_2$- or $\|\cdot\|$-norm (which is $\|\cdot\|_1$-norm in the sparse regression setting),
and {\em $\epsilon$-risk} of recovery---the smallest maximal over $x_*\in X$ radius of $(1-\epsilon)$-confidence ball of norm $|\cdot|$ centered at $\wh x$:
\be
\risk_{|\cdot|,\epsilon}(\wh x|X)=\inf \left\{r:\sup_{x_*\in X}\Prob\{|\wh x-x_*|\geq r\}\leq \epsilon\right\}
\ee{risk2}
\item{\em Prediction risks:} maximal over $x_*\in X$ expected suboptimality
\be
\risk_{g}(\wh x|X)=\sup_{x_*\in X}\bE\{g(\wh x)\}-g_*,
\ee{riskg1}
of $\wh x$ and the smallest maximal over $x_*\in X$ $(1-\epsilon)$-confidence interval 
\be
\risk_{g,\epsilon}(\wh x|X)=\inf \left\{r:\sup_{x_*\in X}\Prob\{g(\wh x)-g_*\geq r\}\leq \epsilon\right\}.
\ee{riskg2}
\end{itemize}
The following statement is a straightforward corollary of Theorems \ref{cor:mycor01} and \ref{cor:reli}.
\begin{proposition}\label{cor:mycorL1} Suppose that \rf{s2bound00} holds. 
\item{(i)} Let the sample size $N$ satisfy
\[N\geq m_0=\left\lceil{16\nu\bar s\over \ul\kappa_\Sigma} (4{e^2}\varkappa\ln[ n]+1)\right\rceil
\] so at least one preliminary stage of Algorithm 1 is completed. Then approximate solutions $\wh x_N$ and $\wh y_N$ produced by the algorithm satisfy
 \begin{eqnarray}\label{d1fina}
  \risk_{\|\cdot\|}(\wh y_N|X)
  &\leq &
  2\sqrt{2s}\risk_{\|\cdot\|_2}(\wh x_N|X)
  \lesssim
 R\exp\left\{-{c N\ul \kappa_\Sigma\over \varkappa\bar s\nu\ln n}\right\}+{\sigma\bar s\over \ul\kappa_\Sigma}\sqrt{\nu\ln n\over  N}\\
   \risk_{g}(\wh x_N|X)
&\lesssim&{\ul\kappa_\Sigma R^2\over \bar s}\exp\left\{-{c N\ul\kappa_\Sigma\over \varkappa\bar s\nu\ln n}\right\}+
{\nu \sigma^2\bar s\varkappa'\ln n\over \ul\kappa_\Sigma N}.\nonumber
 \end{eqnarray}
\item{(ii)} Furthermore, when observation size satisfies $N\geq \alpha m_0\ln[1/\epsilon]$ with large enough absolute $\alpha>0$, $1-\epsilon$ reliable solutions $\wh y_{N,1-\epsilon}$ and $\wh x_{N,1-\epsilon}$ as defined in Section \ref{sec:reliab} satisfy
   \be
   \risk_{\|\cdot\|,\epsilon}(\wh y_{N,1-\epsilon}|X)&\leq& \sqrt{2s}\risk_{\|\cdot\|_2,\epsilon}(\wh y_{N,1-\epsilon}|X)\leq 2\sqrt{2s}\risk_{\|\cdot\|_2,\epsilon}(\wh x_{N,1-\epsilon}|X)\nn
   &\lesssim&
R\exp\left\{-{c N\ul\kappa_\Sigma\over \varkappa\bar s\nu\ln[1/\epsilon]\ln n }\right\}+{\sigma\bar s\over \ul\kappa_\Sigma}\sqrt{\nu\ln[1/\epsilon]\ln n\over  N},
\ee{linhi}
with $\wh x'_{N,1-\epsilon}$, $\wh x''_{N,1-\epsilon}$ and $\wh y'_{N,1-\epsilon}$, $\wh y''_{N,1-\epsilon}$ verifying similar bounds.
\end{proposition}
Let $\sigma_1(\Sigma)$ be the principal eigenvalue (the spectral norm) of $\Sigma$. Then, for all $z$ such that$\|z\|_2=1$ one has
\bse
z^T(\nabla g(x)-\nabla g(x'))&=&\bE\big\{z^T\phi(\myu(\phi^Tx)-\myu(\phi^Tx'))\big\}\leq \ol\bE\big\{|z^T\phi|\,|\phi^T(x-x')|\big\}\\
&\leq& \ol \bE\{(\phi^Tz)^2\}^{1/2}\bE\big\{(\phi^T(x-x'))^2\big\}^{1/2}
\leq 
\ol \sigma_1(\Sigma)\|x-x'\|_2,
\ese
implying that the Lipschitz constant of $\nabla g$ with respect to the Euclidean norm can be set as $\L_2=\ol \sigma_1(\Sigma)$.
Thus, Assumption {\bf S3}  holds when
for some  $1\leq \chi<\infty$ and all $x\in X,\,z\in \bR^n$
\be
{\ul^{-1}\ol}\,\bE\left\{\big(z^T\phi\big)^2\big(\phi^T(x-x_*)\big)^2\right\}\leq \half \chi\|z\|^2_2\sigma_1(\Sigma)\|x-x_*\|_\Sigma^2.
\ee{s3boundL}
Indeed, in this case one has for all $z\in \bR^n$:
\bse
\bE\left\{(z^T\zeta(x,\omega))^2\right\}&=&\bE\left\{
(z^T\phi)^2\left[(\myu(\phi^Tx)-\bE\{\myu(\phi^Tx)\})-(\myu(\phi^Tx_*)-\bE\{\myu(\phi^Tx_*)\})-\sigma\xi\right]^2\right\}\\
&\leq&\bE\left\{\big(z^T\phi\big)^2\big(\myu(\phi^Tx)-\myu(\phi^Tx_*)\big)^2\right\}+\sigma^2\bE\{\xi^2(\phi^Tz)^2\}\\
&\leq&\ol^2\bE\left\{\big(z^T\phi\big)^2\big(\phi^T(x-x_*)\big)^2\right\}
+\sigma^2\|z\|_2^2\sigma_1(\Sigma)\\
\mbox{[by \rf{s3boundL}]}&\leq& \half {\ol\ul} \|x-x_*\|_\Sigma^2\chi\|z\|^2_2\sigma_1(\Sigma)+\sigma^2\|z\|_2^2\sigma_1(\Sigma)\\
&\leq &
(g(x)-g_*)\chi\|z\|^2_2\L_2+\sigma_1(\Sigma)\sigma^2\|z\|_2^2
\ese
implying \rf{s3bound0} with $\chi'={\sigma_1(\Sigma)/ \mynu^2}$ .
\par The following result is a corollary of Theorem \ref{cor:reli2}.
\begin{proposition}\label{cor:mycorL2} Suppose that \rf{s2bound00} and \rf{s3boundL}  hold true, and let
\[
N\geq c\max\left\{{\varkappa\nu\bar s\over \ul\kappa_\Sigma}\ln [1/\epsilon]\ln n,\,{\chi \sigma_1(\Sigma)\over \ul\kappa_\Sigma}\ln[1/\epsilon]
\right\}\]
 with large enough $c>0$. Then aggregated solution $\overline x_{2N,1-\epsilon}$ (with $K=N$) by Algorithm 2 satisfies
   \be
   \risk_{g,\epsilon}(\overline  x_{2N,1-\epsilon}|X)\lesssim {\ul\kappa_{\Sigma}R^2\over \bar s}\exp\left\{-{c N\ul\kappa_\Sigma\over \varkappa\bar s\nu\ln[1/\epsilon]\ln n }\right\}+{\sigma^2\nu\bar s\ln[1/\epsilon]\ln n\over \ul\kappa_\Sigma  N}.
   \ee{linhid}
\end{proposition}
Note that when $\sigma_1(\Sigma)=O(\nu\ln n)$ and $\varkappa$ and $ \chi$ are both $O(1)$ bounds \rf{linhi} and \rf{linhid} hold for $N\geq c{\nu\bar s\over \ul\kappa_\Sigma}\ln [1/\epsilon]\ln n$.
\paragraph{Remark.} Results of Propositions \ref{cor:mycorL1} and \ref{cor:mycorL2} merit some comments. If compared to now standard accuracy bounds for sparse recovery by $\ell_1$-minimization
\cite{bickel2009simultaneous,CT4,CT5,juditsky2011accuracy,raskutti2010restricted,rudelson2012reconstruction,van2009conditions,
candes2011probabilistic}, to the best of our knowledge, \rf{s2bound00} and \rf{s3boundL} provide the most relaxed conditions under which the bounds such as \rf{d1fina}--\rf{linhid} can be established. An attentive reader will notice a degradation of bounds \rf{linhi} and \rf{linhid} with respect to comparable results \cite{dalalyan2019outlier,juditsky2011accuracy,raskutti2010restricted} as far as dependence in factors which are logarithmic in $n$ and $\epsilon^{-1}$ is concerned---bound \rf{yhaterisk} depends on the product $\ln[n]\ln[1/\epsilon]$ of these terms instead of the sum $\ln[n]+\ln[\epsilon^{-1}]$ in the ``classical'' results.\footnote{Note that a similar deterioration was noticed in \cite{candes2011probabilistic}. }  This seems to be a technical ``artifact'' of the analysis of non-Euclidean stochastic approximation algorithm and the reliability {enhancement} approach using median of estimators we have adopted in this work, cf. the comment after Theorem \ref{cor:reli}. Nevertheless,
it is rather surprising to see that conditions on the regressor model in Proposition \ref{cor:mycorL1}, apart from positive definiteness of regressor covariance matrix, essentially amount to (cf. \rf{s2bound000})
\[
\bE\big\{\|\phi\|_\infty^2(\phi^Tz)^2\big\}\lesssim\nu \|z\|_\Sigma^2\;\;\forall z\in \bR^n.
\]
Below we consider some examples of situations where bounds \rf{s2bound000} and \rf{s3boundL} hold with constants which are ``almost'' dimension-independent, i.e. are, at most, {\em logarithmic in problem dimension.} When this is the case, and when observation {count} $N$ satisfies $N\geq \alpha m_0\ln[1/\epsilon]\ln[R/(s\sigma)]$ for large enough absolute $\alpha$, so that the preliminary phase of the algorithm is completed, the bounds of Propositions \ref{cor:mycorL1} and \ref{cor:mycorL2} coincide (up to {already} mentioned logarithmic in $n$ and $1/\epsilon$ factors) with the best accuracy bound available for sparse recovery in the situation in question.\footnote{In the case of ``isotropic sub-Gaussian'' regressors, see \cite{lecue2018regularization}, the bounds of Proposition \ref{cor:mycorL1} are comparable to bounds of \cite[Theorem 5]{lecue2017robust} for Lasso recovery under relaxed moment assumptions on the noise $\xi$.}
 \begin{enumerate}
\item {\em Sub-Gaussian regressors:} suppose now that $\phi_i\sim\SG(0,S)$, i.e., regressors $\phi_i$ are sub-Gaussian with zero mean and matrix parameter $S$, meaning that
     \[\bE\big\{e^{u^T\phi}\big\}\leq e^{u^TS u \over 2 } \;\;\mbox{for all $u\in\bR^n$.}\]
      Let us assume that sub-Gaussianity matrix $S$ is ``similar'' to the covariance matrix $\Sigma$ of $\phi$, i.e.  $S\preceq \mu\Sigma$ with some $\mu<\infty$. Note that $\bE\{(\phi^Tz)^4\}\leq 16(z^TSz)^2\leq 16\mu^2\|z\|_\Sigma^4$, and thus
\[
\bE\{(z^T\phi\phi^Tx)^2\}\leq \bE\{(z^T\phi)^4\}^{1/2}\bE\{(x^T\phi)^4\}^{1/2}\leq 16 z^TSz\,x^TSx\leq 16\mu^2\sigma_1(\Sigma)\|z\|^2_2\|x\|_\Sigma^2,
\]
which is \rf{s3boundL} with $\chi=16\mu^2\ul^{-1}\ol$.  Let us put $\bar\upsilon=\max_i[S]_{ii}$. One easily verifies that in this case
\[
\mynu^2=\bE\{\|\phi\|_\infty^2\}\leq 2\bar\upsilon(\ln [2n]+1)\leq 2\mu\upsilon(\ln [2n]+1),
\]
and
\[
\bE\{\|\phi\|_\infty^4\}\leq 4\bar \upsilon^2(\ln^2[2n]+2\ln [2n]+2)\leq
4\mu^2\upsilon^2(\ln^2[2n]+2\ln [2n]+2).
\]
As a result, we have, cf. \rf{s2bound00},
\bse
\varsigma^2(x)
&\leq &\left[\ol(\bE\{\|\phi\|_\infty^4\})^{1/4}(\bE\{(\phi^T(x-x_*))^4\})^{1/4}+\sigma (\bE\{\|\phi\|_\infty^2\})^{1/2}+\ol\sqrt{\upsilon}\|x-x_*\|_{\Sigma}\right]^2\\
&\leq &\left[\ol\sqrt{8\bar \upsilon(\ln [2n]+2)}\|x-x_*\|_S+\sigma \sqrt{2\bar \upsilon(\ln [2n]+1)}+\ol\sqrt{\upsilon}\|x-x_*\|_{\Sigma}\right]^2\\
&\leq& 2\ol^2\big(\mu\sqrt{8(\ln [2n]+2)}+1\big)^2\upsilon\|x-x_*\|^2_\Sigma+4\mu \upsilon(\ln [2n]+1)\sigma^2,
\ese
whence, Assumption {\bf S1} holds with $\varkappa\nu\lesssim \ol^2\ul^{-1}\mu^2\upsilon\ln n$,
$\varkappa'\lesssim 1$, and $\varsigma_*^2\lesssim\mu\upsilon\sigma^2\ln n$.
\item {\em Bounded regressors:} we assume that $\|\phi_i\|_\infty\leq \mu$ a.s.. One has
\bse
\varsigma^2(x)&\leq & \left(\ol\mu\bE\{(\phi^T(x-x_*))^2\}^{1/2}+\ol\upsilon^{1/2} \|x-x_*\|_\Sigma+\mu\sigma\right)^2\\
&\leq&
2\ol^2(\mu+\upsilon^{1/2})^2\|x-x_*\|_{\Sigma}^2+2\mu^2\sigma^2,
\ese
implying the second inequality of \rf{s2bound00} and also \rf{s2bound0} with $\varkappa\nu\leq 4\ol^2\ul^{-1}(\mu+\sqrt{\upsilon})^2$ and $\varsigma_*^2\leq \mu^2\sigma^2$.
In particular, this condition is straightforwardly satisfied when $\phi_j$ are sampled from an orthogonal system with uniformly bounded elements, e.g., $\phi_j=\sqrt{n}\psi_{\kappa_j}$ where $\{\psi_j,j=1,...,n\}$ is a trigonometric or Hadamard basis of $\bR^n$, and $\kappa_j$ are independent and uniformly distributed over $\{1,...,n\}$. On the other hand, in the latter case, for  $z=x=\psi_1$ we have
\[
\bE\{(z^T\phi\phi^Tx)^2\}=\bE\{(\psi_1\phi\phi^T\psi_1)^2\}=n=n\|\psi_1\|^4_2=n\|x\|_2^2\|z\|_2^2,
\]
implying that \rf{s3boundL} can only hold with $\chi=O(n)$ in this case.

%
\par
Besides this, when $ \phi$ is a linear image of a Rademacher vector, i.e. $\phi=A\eta$ where $A\in \bR^{m\times n}$ and $\eta$ has independent components $[\eta]_i\in \{\pm 1\}$ with $\Prob\{[\eta]_i=1\}=\Prob\{[\eta]_i=-1\}=1/2$, one has $\Sigma=AA^T$, and $\bE\{(\phi^Tx)^4\}\leq 3\|A^Tx\|_2^4$ (cf. the case of sub-Gaussian regressors above). Thus, we have
\bse
\bE\{(z^T\phi\phi^T(x-x_*))^2\}&\leq& \bE\{(z^T\phi)^4\}^{1/2}\bE\{((x-x_*)^T\phi)^4\}^{1/2}\\&\leq& 3z^T\Sigma z\,(x-x_*)^T\Sigma (x-x_*)\leq 3\sigma_1(\Sigma)\|z\|_2^2\|x-x_*\|_\Sigma^2
\ese
implying \rf{s3boundL} with $\chi=6\ul^{-1}\ol$. On the other hand, when denoting $\mu=\max_{j}\|\Row_j(A)\|_2$, we get  $\Prob\{\|\phi\|_\infty^4\geq t\mu\}\leq 2ne^{-t^2/2}$ with
\[
\bE\{\|\phi\|_\infty^2\}\leq 2\mu^2[\ln[2n]+1] \;\;\mbox{and}\;\;\bE\{\|\phi\|_\infty^4\}\leq 4\mu^4[\ln^2[2n]+2\ln[2n]+2 ].
\]
Thus, by \rf{s2bound00},
\bse
\zeta^2(x)
&\leq &\left(\ol(\bE\{\|\phi\|_\infty^4\})^{1/4}(\bE\{(\phi^T(x-x_*))^4\})^{1/4}+\sigma (\bE\{\|\phi\|_\infty^2\})^{1/2}+\ol\sqrt{\upsilon}\|x-x_*\|_{\Sigma}\right)^2\\
&\leq &\left(\ol\sqrt{2\sqrt{3}(\ln [2n]+2)}\mu\|x-x_*\|_\Sigma+\sigma \sqrt{2(\ln [2n]+1)}\mu+\ol\sqrt{\upsilon}\|x-x_*\|_{\Sigma}\right)^2\\
&\leq &
2\mu^2\ol^2\big(\sqrt{2\sqrt{3}(\ln [2n]+2)}+1\big)^2\|x-x_*\|^2_\Sigma+4{\mu^2} (\ln [2n]+1)\sigma^2\ese
which is \rf{s2bound0} with $\varkappa\nu\lesssim \mu^2\ul^{-1}\ol^2\ln n$ and
$\varkappa'\varsigma_*^2\lesssim\mu^2\sigma^2\ln n$.
\item{\em Scale mixtures:}
Let us now assume that
\begin{equation}\label{zeq4}
\phi\sim \sqrt{Z}\eta,
\end{equation}
where $Z$ is a scalar a.s. positive random variable, and $\eta\in \bR^n$ is independent of $Z$ with covariance matrix  $\bE\{\eta\eta^T\}=\Sigma_0$. Because
  \[
  \bE\big\{\|\phi\|_\infty^2\big\}=\bE\{Z\}\bE\big\{\|\eta\|_\infty^2\big\},\;\;\bE\big\{\|\phi\phi^Tz\|_\infty^2\big\}=\bE\{Z^2\}\bE\big\{\|\eta\eta^Tz\|_\infty^2\big\}
   \]
and
  \[
  [\Sigma:=]\;\bE\{\phi\phi^T\}=\bE\{Z\}\bE\{\eta\eta^T\},
   \]
we conclude that if  random vector $\eta$ satisfies \rf{s2bound00} with $\Sigma_0$ substituted for $\Sigma$ and $\bE\{Z^2\}$ is finite then a similar bound also holds for $\phi$. It is obvious that if $\eta$ satisfies \rf{s3boundL} then
  \[
  \bE\big\{(z^T\phi\phi^Tx)^2\big\}=\bE\{Z^2\}\bE\big\{(z^T\eta\eta^Tx)^2\big\}\leq {\bE\{Z^2\}\over \bE\{Z\}^2}\chi\|z\|_{\Sigma}^2\|x\|^2_{\Sigma}\leq\chi{\bE\{Z^2\}\over \bE\{Z\}^2}\sigma_1(\Sigma)\|z\|^2_2\|x\|_\Sigma^2,
   \]
   and \rf{s3boundL} holds for $\phi$ with $\chi$ for $\eta$ replaced with $\chi{\bE\{Z^2\}\over \bE\{Z\}^2}$.
\par
Let us consider the situation where $\eta\sim\N(0,\Sigma_0)$ with positive definite $\Sigma_0$. In this case $\phi$ is referred to as Gaussian scale mixture with a standard example provided by {\em $n$-variate  $t$-distributions} $t_n(q,\Sigma_0)$ (multivariate Student distributions with $q$ degrees of freedom, see \cite{kotz2004multivariate} and references therein). Here, by definition, $t_n(q,\Sigma_0)$ is the distribution of the random vector
  $\phi=\sqrt{Z}\eta$ with $Z=q/\zeta$, where $\zeta$ is the independent of $\eta$ $\chi^2$-random variable with $q$ degrees of freedom. One can easily see that all one-dimensional projections $e^T\phi$, $\|e\|_2=1$, of $\phi$ are random variables with univariate $t_q$-distribution. When $\phi_i\sim t_n(q,\Sigma_0)$ with $q>4$, we have for $\zeta\sim \chi^2_q$
  \[
  \bE\left\{{q\over \zeta}\right\}={q\over q-2},\;\;\;\bE\left\{{q^2\over \zeta^2}\right\}={3q^2\over (q-2)(q-4)},
  \] so that $\Sigma= {q\over q-2}\Sigma_0$, and
  \bse
\varsigma^2(x)
 &\lesssim &\ol^2
 {q-2\over q-4}\upsilon\ln[n] \|x-x_*\|_\Sigma+\sigma^2\upsilon\ln n
\ese
implying \rf{s2bound0} with $\varkappa\nu \lesssim\ol^2\ul^{-1} \upsilon\ln n,\,\varkappa'\lesssim 1$, and $\varsigma_*^2\lesssim\sigma^2\upsilon\ln n$.
Moreover, in this case
\[
\bE\{(z^T\phi\phi^Tx)^2\}=\bE\{Z^2\}\bE\{z^T\eta\eta^Tx)^2\}\leq 3{\bE\{Z^2\}\over \bE\{Z\}^2}\|z\|^2_\Sigma\|x\|^2_\Sigma\leq 9 {q-2\over q-4} \sigma_1(\Sigma)\|z\|^2_2\|x\|^2_\Sigma.
\]
 Another example of Gaussian scale mixture \rf{zeq4}
 is the {\em $n$-variate Laplace distribution} $\L_n(\lambda,\Sigma_0)$ \cite{eltoft2006multivariate} in which $Z$ has exponential distribution with parameter $\lambda$. In this case all one-dimensional projections $e^T\phi$, $\|e\|_2=1$, of $\phi$ are Laplace random variables. If $\phi_i\sim \L_n(\lambda,\Sigma_0)$ one has
  \[
\varsigma^2(x)\lesssim  \ol^2\upsilon\ln[ n] \|x-x_*\|_\Sigma+\sigma^2\upsilon\ln n
\]
and
\[\bE\{(z^T\phi\phi^Tx)^2\}\lesssim\sigma_1(\Sigma)\|z\|^2_2\|x\|^2_\Sigma.
\]
 \end{enumerate}
\subsection{Stochastic Mirror Descent for low-rank matrix recovery}
\label{sec:matrec}
In this section we consider the problem of recovery of matrix~$x_{*}\in \mathbf{R}^{p\times q}$, from independent and identically distributed observations
\be
\eta_i=\left\langle \phi_i, x_* \right\rangle+\sigma\xi_i,\;\;\;i=1,2,...,N,
\ee{lrobs}
with $\phi_i\in \mathbf{R}^{p\times q}$ which are random independent over $i$ with covariance operator
$\Sigma$ (defined according to $\Sigma(x)=\bE\{\phi\la\phi,x\ra\}$). We assume that   $\xi_i\in \mathbf{R}$ are mutually independent and independent of $\phi_i$ with
$\bE\{\xi_i\}=0$ and $\bE\{\xi_i^2\}\leq 1$.
\par
In this application, $E$ is the space of $p\times q$ matrices equipped with the Frobenius scalar product
\[
    \left\langle a, b\right\rangle = \text{Tr}\left( a^T b\right)
\]
with the corresponding norm $\|a\|_2=\la a,a\ra^{1/2}$. For the sake of definiteness, we assume that $p\geq q\geq 2$. Our choice for the norm $\|\cdot\|$ is the nuclear norm $\|x\|=\|\sigma(x)\|_1$ where $\sigma(x)$ is the singular spectrum of $x$, so that the conjugate norm is the spectral norm $\|y\|_*=\|\sigma(y)\|_\infty$.
We suppose that
\[
\kappa_\Sigma \|x\|^2_2\leq \la x,\Sigma(x)\ra \leq \upsilon \|x\|^2_2 \;\forall x\in \bR^{p\times q},
\]
with known $\kappa_\Sigma>0$ and $\upsilon$, we write $\kappa_\Sigma I\preceq \Sigma\preceq\upsilon I$; for $x\in \bR^{p\times q}$ we denote $\left\|x \right\|_{\Sigma} = \sqrt{\la x, \Sigma(x)\ra}$.
Finally, we assume that matrix $x_*$ is of rank $s\leq \bar s\leq q$, and moreover, that we are given a convex and closed subset $X$ of $\bR^{p\times q}$ such that $x_*\in X$, along with $R<\infty$ and $x_0\in X$ satisfying $\|x_*-x_0\|\leq R$.

\par Consider the Stochastic Optimization problem
\be
\min_{x\in X}\left\{g(x)=\half \bE\big\{\underbrace{(\eta-\left\langle\phi, x\right\rangle)^2}_{=:G(x,\omega=[\phi,\eta])}\big\}\right\}.
\ee{SOP.lr}
We are to apply SMD algorithm to solve \rf{SOP.lr} with the proximal setup associated with  the nuclear norm with  quadratically growing  for $q\geq 2$ distance-generating function
\[
\vartheta(x)=2e\ln(2q)\left[\sum_{j=1}^q \sigma^{1+r}_j(x)\right]^{2\over 1+r},\;\;r=\big(1 2\ln [2q]\big)^{-1}
\]
(here $\sigma_j(x)$ are singular values of $x$), with the corresponding parameter $\Theta\leq C \ln[2q]$ (cf. \cite[Theorem 2.3]{nesterov2013first}).
Note that, in the premise of this section,
\[
g(x)=\half \bE\big\{(\sigma\xi+\la \phi,x_*-x\ra)^2\big\}=\half (\|x-x_*\|_\Sigma^2+\sigma^2),
\]
with
\[\nabla g(x)=\Sigma(x-x_*)=\bE\big\{\underbrace{\phi(\la \phi,x-x_*\ra-\sigma\xi)}_{=\nabla G(x,\omega)}\big\}
\]
and
\[
\zeta(x,\omega)=\nabla G(x,\omega)-\nabla g(z)=[\phi\la \phi,x-x_*\ra -\Sigma(x-x_*)]-\sigma \phi\xi.\]
%
%

Let us now consider the case regressors $\phi_i\in \mathbf{R}^{p\times q}$ drawn independently from a {\em sub-Gaussian ensemble}, $\phi_i \sim \SG(0,S)$ with sub-Gaussian operator $S$. The latter means that
\[
\bE\big\{e^{\la x,\phi\ra}\big\}\leq e^{\half \la x,S(x)\ra }\;\;\forall x\in \bR^{p\times q}
\]
with linear positive definite $S(\cdot)$.
To show the bound of Theorems \ref{cor:mycor01}--\ref{cor:reli2} in this case we need to verify that relationships \rf{s2bound0} and \rf{s3bound0} of Assumptions {\bf S1} and {\bf S3} are satisfied.
To this end, let us assume that $S$ is ``similar'' to the covariance operator $\Sigma$ of $\phi$, namely, $S\preceq \mu\Sigma$ with some $\mu<\infty$. This setting covers, for instance, the situation where the entries in the regressors matrix $\phi\in \mathbf{R}^{p\times q}$ are standard Gaussian or Rademacher  i.i.d. random variables (in these models, $S=\Sigma$ is the identity, and $g(x) - g_* = \half \|x-x_*\|_2^2$).

Note that, more generally, when $S\preceq \mu\Sigma$ we have $S\preceq \mu \upsilon I$ with
\[
\bE\{\|\phi\|_*^4\}\leq C^2\mu^2\upsilon^2 (p+q)^2,
\]
cf. Lemma \ref{lem:spectral.conc} of the appendix, and
\[
\bE\{\la\phi,x-x_*\ra^4\}\leq 16\la x-x_*,S(x-x_*)\ra^2\leq 16\mu^2\|x-x_*\|_\Sigma^2
\]
for sub-Gaussian  random variable $\la\phi,x-x_*\ra\sim \SG(0,\la x-x_*, S(x-x_*)\ra)$. Therefore,
\bse
     \lefteqn{\bE\big\{\|\phi\la \phi,x-x_*\ra-\Sigma(x-x_*)\|_*^2\big\}\leq2\bE\big\{\|\phi\la \phi,x-x_*\ra\|_*^2\big\}+2\upsilon\|x-x_*\|^2_\Sigma}\\
    &\leq&
    2\bE\big\{\|\phi\|_*^4\big\}^{1/2}\bE\big\{\la\phi,x-x_*\ra^4\big\}^{1/2}
    +2\upsilon\|x-x_*\|^2_\Sigma\\
    &\leq&
    8C\mu^2 (p+q)\upsilon \|x-x_*\|_\Sigma^2+2\upsilon\|x-x_*\|^2_\Sigma.
\ese
 Taking into account that $\nu=\bE\{\|\phi\|^2_*\}\leq C\mu\upsilon (p+q)$ in this case, we have
\bse
\varsigma^2(x)&=&\bE\big\{\|\zeta(x,\omega)\|_*^2\big\}\leq 2\bE\big\{\|\phi\la \phi,x-x_*\ra -\Sigma(x-x_*)\|^2_*\big\}+
{2}\sigma^2 \bE\big\{\|\phi\|_*^2\big\}\\
&\leq& 8(4C\mu^2(p+q)+1)\upsilon[g(x)-g_*]
+2\underbrace{C\mu\upsilon (p+q)\sigma^2}_{=\varsigma^2_*}
\ese
implying \rf{s2bound0} with $\varkappa \lesssim \mu$ and $\varkappa'\lesssim 1$.
\par
Similarly, we estimate $\forall x\in X,\,z\in \bR^{p\times q}$
\[
\bE\left\{
\left\langle \phi,z\right\rangle^2
\left\langle \phi,x\right\rangle^2
\right\}
\leq
\bE\left\{
\left\langle z,\phi\right\rangle^4
\right\}^{1/2}
\bE\left\{
\left\langle \phi,x\right\rangle^4
\right\}^{1/2}
\leq
16 \la z, S(z)\ra \la x, S(x)\ra \leq 16\mu^2\upsilon\|z\|_2^2\|x\|_\Sigma^2,
\]
so that
\bse
\bE\left\{\la z,\zeta(x,\omega)\ra^2\right\}
&=&\bE\left\{
\left\langle z,\phi\la \phi,x-x_*\ra-\Sigma(x-x_*)-\sigma\phi\xi\right\rangle^2\right\}\nn
&=&\bE\left\{\big(\la z,\phi\ra \la \phi,x-x_*\ra -\la z,\Sigma(x-x_*)\ra\big)^2\right\}
+\sigma^2\bE\{\xi^2\la z,\phi\ra^2\}\nn
&\leq&\bE\left\{\la z,\phi\ra^2
\la \phi, x-x_*\ra^2\right\}+\sigma^2\upsilon\|z\|_2^2\nn
&\leq& 16\mu^2\upsilon[g(x)-g_*]\|z\|^2_2+\sigma^2\upsilon\|z\|_2^2
\ese
implying the bound \rf{s3bound0} with $\chi\lesssim \mu(p+q)^{-1}$ and $\chi'\lesssim {\mu^{-1}}(p+q)^{-1}$.
When substituting the above bounds for problem parameters into statements of Theorems \ref{cor:mycor01}--\ref{cor:reli2} we obtain the following statement summarizing the  properties of the approximate solutions by the SMD-SR algorithm utilizing observations \rf{lrobs}; the corresponding risks are defined in \rf{risk1}--\rf{riskg2}.
\begin{proposition}\label{cor:mycorLR} In the situation of this section,
\item{(i)} let the sample size $N$ satisfy
\[N\geq \alpha\left[{\mu^2\upsilon(p+q)\bar s\ln q\over \kappa_\Sigma} \right]
\] for an appropriate absolute $\alpha$, implying that at least one preliminary stage of Algorithm 1 is completed. Then there is an absolute $c>0$ such that approximate solutions $\wh x_N$ and $\wh y_N$ produced by the algorithm satisfy
 \bse
  \risk_{\|\cdot\|}(\wh y_N|X)
  &\leq &
  2\sqrt{2s}\risk_{\|\cdot\|_2}(\wh x_N|X)
  \lesssim
 R\exp\left\{-{c N\kappa_\Sigma\over \mu^2\upsilon (p+q)\bar s\ln q}\right\}
 +{\sigma\bar s\over \kappa_\Sigma}\sqrt{\mu\upsilon(p+q)\ln q\over  N},\\
   \risk_{g}(\wh x_N|X)
&\lesssim&{\kappa_\Sigma R^2\over \bar s}\exp\left\{-{c N\kappa_\Sigma\over \mu^2\upsilon (p+q)\bar s\ln q}\right\}+
{\sigma^2\mu\upsilon(p+q)\bar s\ln q\over \kappa_\Sigma N}.
 \ese
\item{(ii)} Furthermore, when observation size satisfies \[N\geq \alpha'\left[{\mu^2\upsilon(p+q)\bar s\ln[1/\epsilon]\ln q\over \kappa_\Sigma} \right]\] with large enough $\alpha'$, $(1-\epsilon)$-reliable solutions $\wh y_{N,1-\epsilon}$ and $\wh x_{N,1-\epsilon}$ defined in Section \ref{sec:reliab} satisfy for some $c'>0$
   \be
   \lefteqn{\risk_{\|\cdot\|,\epsilon}(\wh y_{N,1-\epsilon}|X)\leq \sqrt{2s}\risk_{\|\cdot\|_2,\epsilon}(\wh y_{N,1-\epsilon}|X)\leq 2\sqrt{2s}\risk_{\|\cdot\|_2,\epsilon}(\wh x_{N,1-\epsilon}|X)}\nn
   &\lesssim&
R\exp\left\{-{c' N\kappa_\Sigma\over \mu^2\upsilon (p+q)\bar s\ln[1/\epsilon]\ln q}\right\}+{\sigma\bar s\over \kappa_\Sigma}\sqrt{\mu\upsilon(p+q)\ln[1/\epsilon]\ln q\over  N},
\ee{linhilr}
with solutions $\wh x'_{N,1-\epsilon}$, $\wh x''_{N,1-\epsilon}$ and $\wh y'_{N,1-\epsilon}$, $\wh y''_{N,1-\epsilon}$ verifying analogous bounds. Finally, the following bound holds for the aggregated solution $\overline x_{2N,1-\epsilon}$  (with $K=N$) by Algorithm 2:
   \bse
   \risk_{g,\epsilon}(\overline  x_{2N,1-\epsilon}|X)\lesssim {\kappa_\Sigma R^2\over \bar s}\exp\left\{-{c' N\kappa_\Sigma\over \mu^2\upsilon (p+q)\bar s\ln[1/\epsilon]\ln q}\right\}+
{\sigma^2\mu\upsilon(p+q)\bar s\ln[1/\epsilon]\ln q\over \kappa_\Sigma N}.
   \ese
\end{proposition}
\paragraph{Remark.} Let us now compare the bounds of the proposition to available accuracy estimates for low rank matrix recovery. Notice first, that when assuming that $\mu\lesssim1$ the bounds of the proposition hold if (the upper bound on unknown) signal rank $\bar s$ satisfies
\[
\bar s\lesssim {N\kappa_\Sigma\over (p+q)\upsilon\ln[1/\epsilon]\ln q}.
\]
The above condition is essentially the same,  up to logarithmic in $1/\epsilon$ factor, as the best condition on rank of the signal to be recovered under which the recovery is exact in the case of exact---noiseless---observation \cite{candes2009tight,recht2010guaranteed}.
The risk bounds of Proposition \ref{cor:mycorLR} can be compared to the corresponding accuracy bounds for recovery $\wh x_{N,\mathrm{Lasso}}$ by Lasso with nuclear norm penalization, as in \cite{negahban2011estimation,koltchinskii2011nuclear}. For instance, when regressors $\phi_i$ have i.i.d. $\N(0,1)$ entries they state (cf. \cite[Corollary 5]{negahban2011estimation}) that the $\|\cdot\|_{2,\epsilon}$-risk of the recovery satisfies the bound
\[
\risk_{\|\cdot\|_2,\epsilon}(\wh x_{N,\mathrm{Lasso}}|X)\lesssim {\sigma^2 r(p+q)\over N}
\]
for $\epsilon\geq \exp\{-(p+q)\}$. Observe that the above bound coincides, up to logarithmic in $q$ and $1/\epsilon$ factors with the second---asymptotic---term in the bound \rf{linhilr}. This result is all the more surprising if we recall that its validity is not limited to sub-Gaussian regressors---what we need in fact is the bound (cf. the remark after Proposition \ref{cor:mycorL2})
\be
\bE\big\{\|\phi\la \phi,z\ra\|_*^2\big\}\lesssim (p+q)\|x-x_*\|_\Sigma^2.
\ee{lrcond}
For instance, one straightforwardly verifies that the latter bound holds, for instance, in the case where regressor $\phi$ is a scale mixtures of matrices satisfying \rf{lrcond} (e.g., scale mixture of sub-Gaussian matrices).

\section{Numerical illustration}
\label{sec.exp}
We present results of a preliminary simulation study illustrating performance of the SMD-SR algorithm.
\paragraph{Experimental setting.}
We present results of simulated experiments of sparse linear regression \rf{sparselin} with linear activation $\myu(t)=t$ and i.i.d. random $(\phi_i,\xi_i)$
in the setting~$N\le n$ with $(n,s) = (100\,000, 50)$.
In our experiments, covariance matrix~$\Sigma$ of regressors is diagonal with diagonal entries~$\Sigma_{11}\le\Sigma_{22}\le\dots\le\Sigma_{nn}$ evenly spaced over $[{\kappa_\Sigma}, \nu]$, parameters~$({\kappa_\Sigma}, \nu)$ being specific for each experiment.
The indices of nonvanishing components of the optimal solution $x_*$ are evenly spaced in $[1,n]$ with the non-zero entries being sampled from the standard Gaussian distribution.
The number~$s$ of nonzero components of $x_*$ and the value $\kappa_\Sigma$ are assumed to be known.

%
We compare the performance of the SMD-SR procedure to that of the ``vanilla'' non-Euclidean SMD algorithm utilizing the same proximal setup when solving stochastic optimization problem \rf{SOP}. Another contender is  the coordinate descent algorithm (CDA) of the Python package {\tt sklearn} solving the Lasso problem
\begin{equation}
    \min_{x\in \bR^n}
    \bigg\{
        \frac1{2N}\sum_{i=1}^{N}
        [\eta_i-\phi_i^Tx]^2 +
        \lambda \|x\|_1
    \bigg\}
    \label{exp.lasso.def}
\end{equation}
with the ``theoretically optimal'' choice $\lambda =
    2\sigma\sqrt{\frac{2\ln n}N}$ of the penalty parameter (cf. \cite{bickel2009simultaneous,koltchinskii2011nuclear}).

\paragraph{Parameter setting for SMD-SR.}
As it is often the case, the theoretical choice of algorithm parameters as given in Sections \ref{sec:multis} and \ref{sec:sr} is too conservative in practice.
We give a brief overview of the workarounds used in our simulations.
\bigskip
\begin{itemize}
\item We use stages of fixed length and mini-batches of exponentially increasing size during the asymptotic phase of the method, cf. \cite[Section 4.5]{andreiphd}.
This allows to significantly accelerate computations at the asymptotic regime alleviating the computational burden of prox-evaluations.
\item We use variable stepsize parameters $\beta_i = \beta_0\|\phi_i\|_\infty^2$ with constant $\beta_0 = 1.0$ both for SMD-SR and SMD.
This choice of $\beta_0$ corresponds to the condition~$\beta_0\ge \nu$ but neglects the constants factors arising in the theoretical analysis.
In order to compute the current approximate solution, the estimates of the SMD algorithm are then weighted with the corresponding~$\beta_i$.
\item The number of steps~$m_0$ to be performed by the SMD algorithm on each stage is set to $m_0=\left\lceil(1/2) s \nu(\ln [n]+1)\right\rceil$, which corresponds to~\rf{betapr} in the case of~${\kappa_\Sigma}=1.0$.
\item
In our simulations, we utilize the CUSUM test for monitoring a change detection~(see, e.g., \cite{ploberger1992cusum,lee2003cusum}) to decide upon switching from preliminary (``linear trend'') to asymptotic phase (``sublinear trend'') of the algorithm; however, we perform at least 4 preliminary stages.
\end{itemize}

\paragraph{Experimental results.}
We present results of two series of experiments, experiments in each series corresponding to 4 combinations of parameters ~${\kappa_\Sigma}$ and $\sigma$ with~${\kappa_\Sigma}\!\in\!\{0.1, 1.0\}$ and~$\sigma\!\in\!\{0.001, 0.1\}$; we run 20 simulations for each parameter combination.
In the figures below, for each ``contender'' we plot the median value of the prediction error $\|\wh x_t-x_*\|_\Sigma$ as a function of $t=1,...,N$ along with the tubes of 25\% and 75\% quantiles.

\begin{figure}[!hptb]
\vspace*{-0.2cm}
\begin{tabular}{ll}
\includegraphics[width=0.45\textwidth]{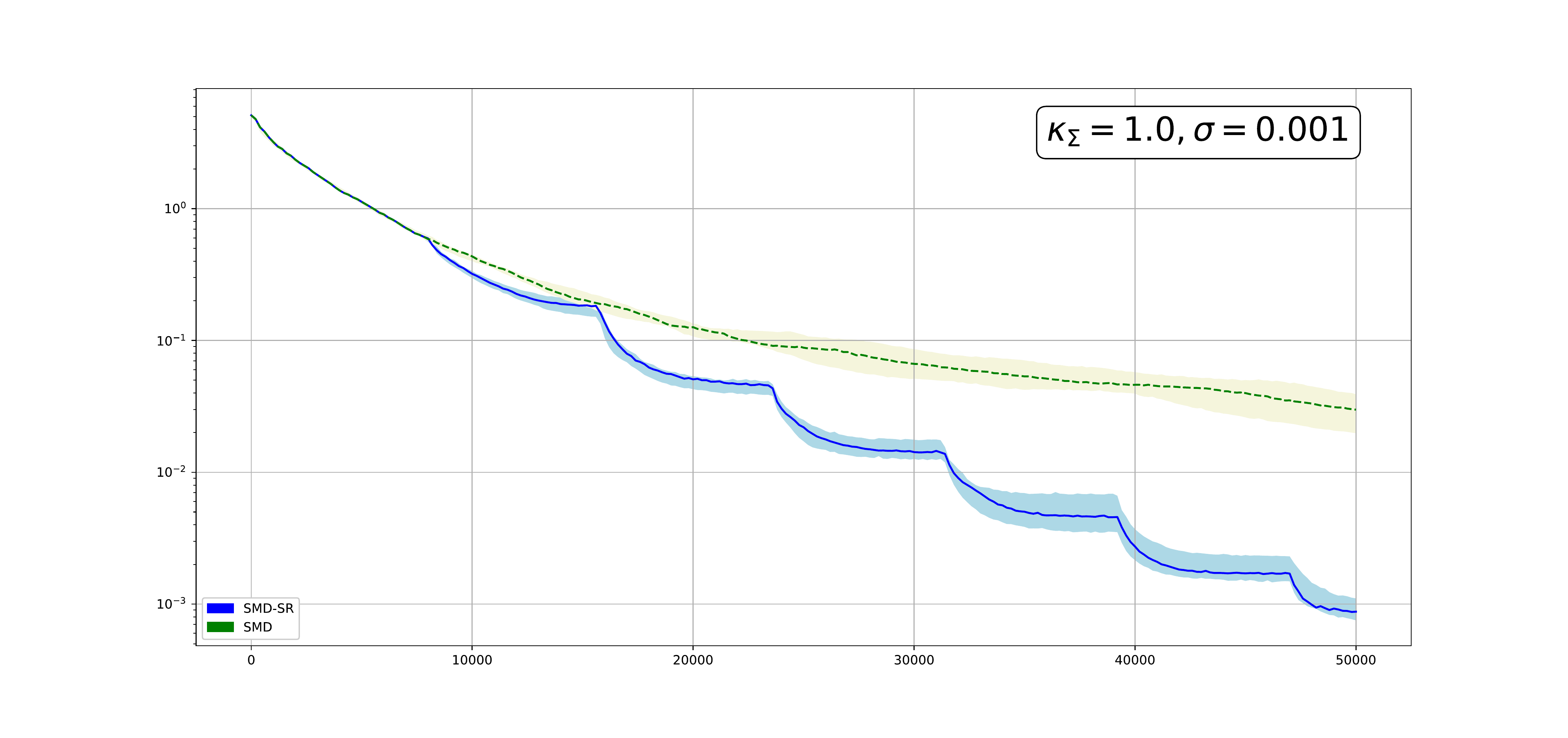}&
\includegraphics[width=0.45\textwidth]{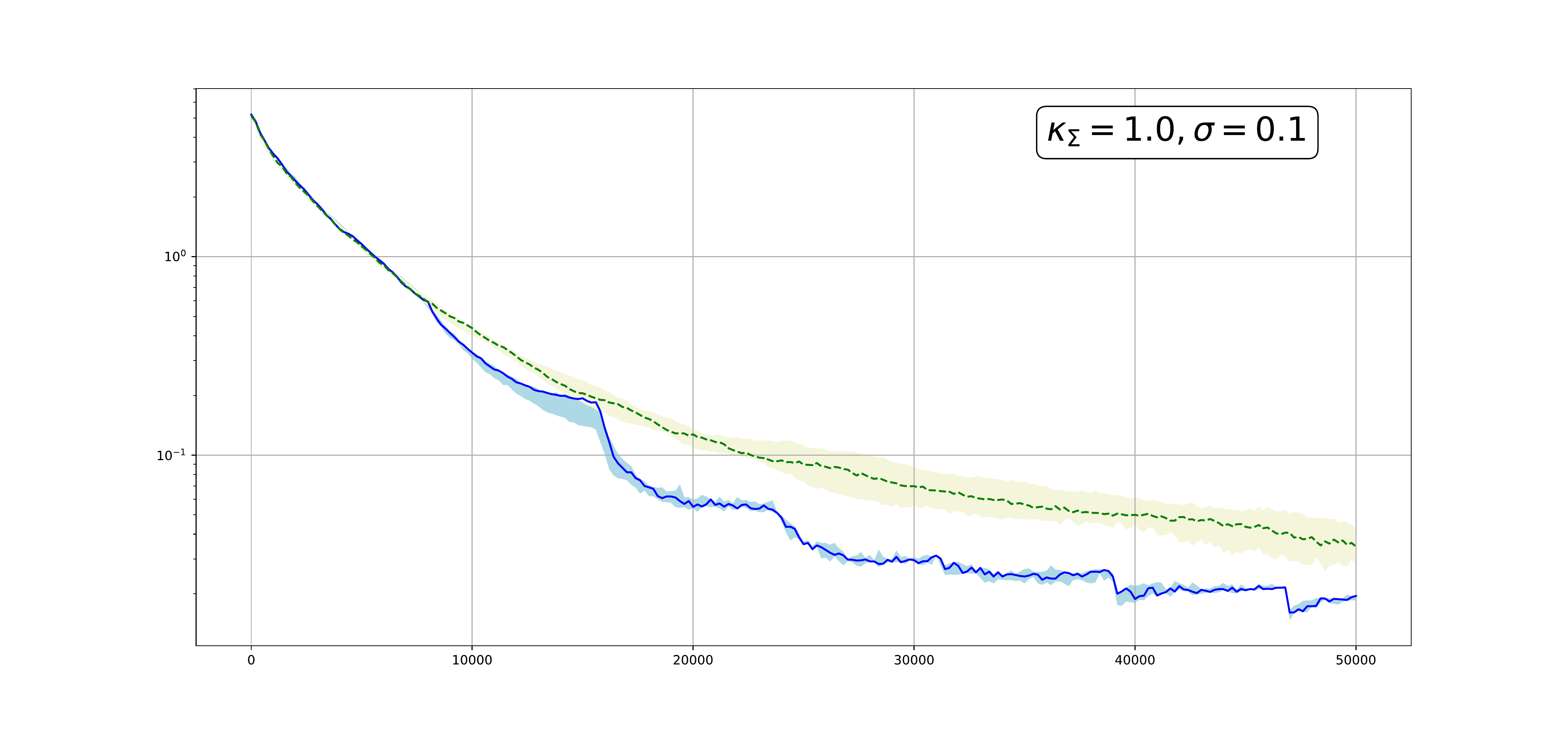}\\
\includegraphics[width=0.45\textwidth]{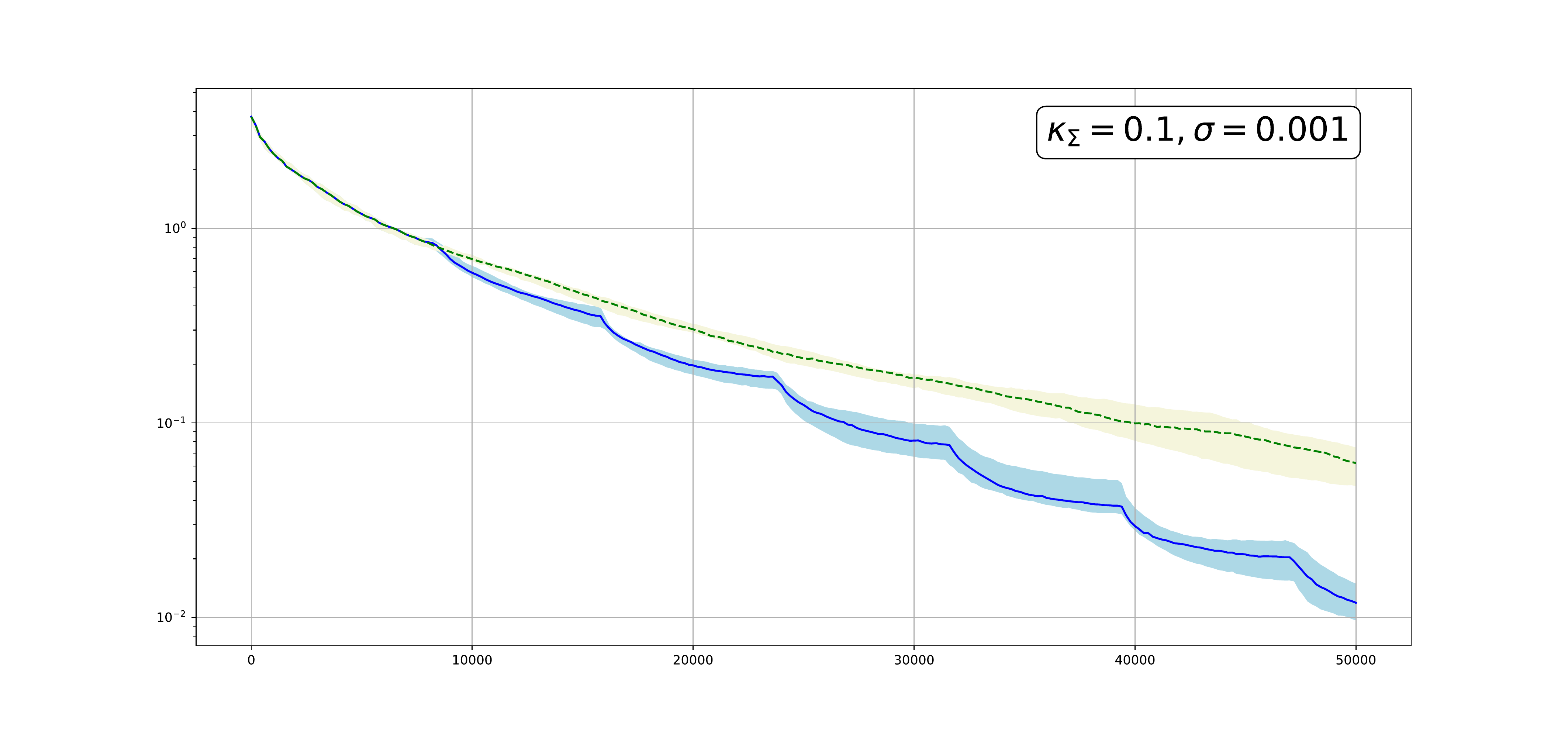}&
\includegraphics[width=0.45\textwidth]{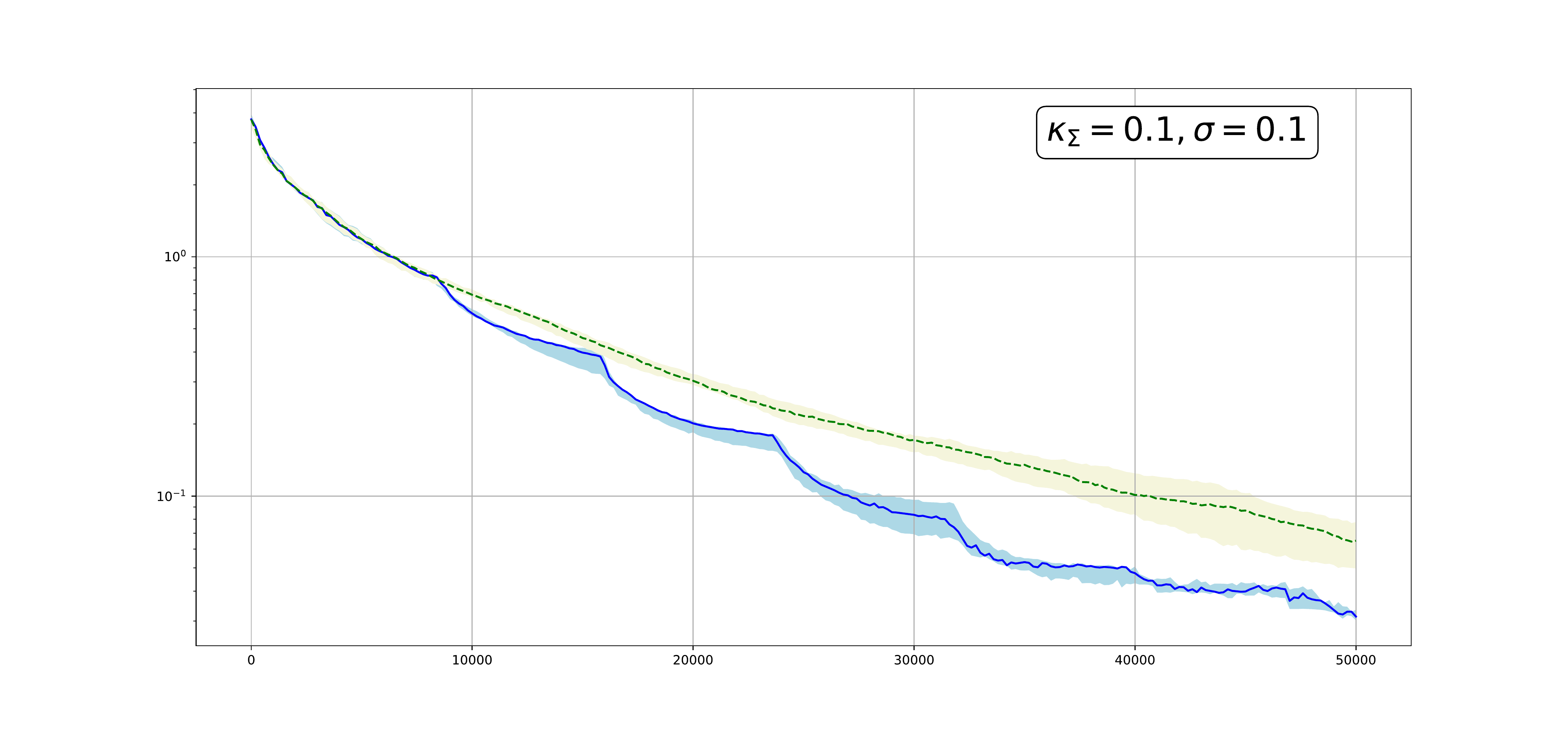}
\end{tabular}
\vspace*{-0.5cm}\caption{Comparison of SMD-SR (solid line) and SMD (dashed line) in the Gaussian setting; $(n,s) = (100\,000,  50)$.
}
\label{fig:normal.50000}
\end{figure}

\begin{center}
\begin{figure}[!h]
\vspace*{-0.5cm}\begin{tabular}{ll}\includegraphics[width=0.45\textwidth]{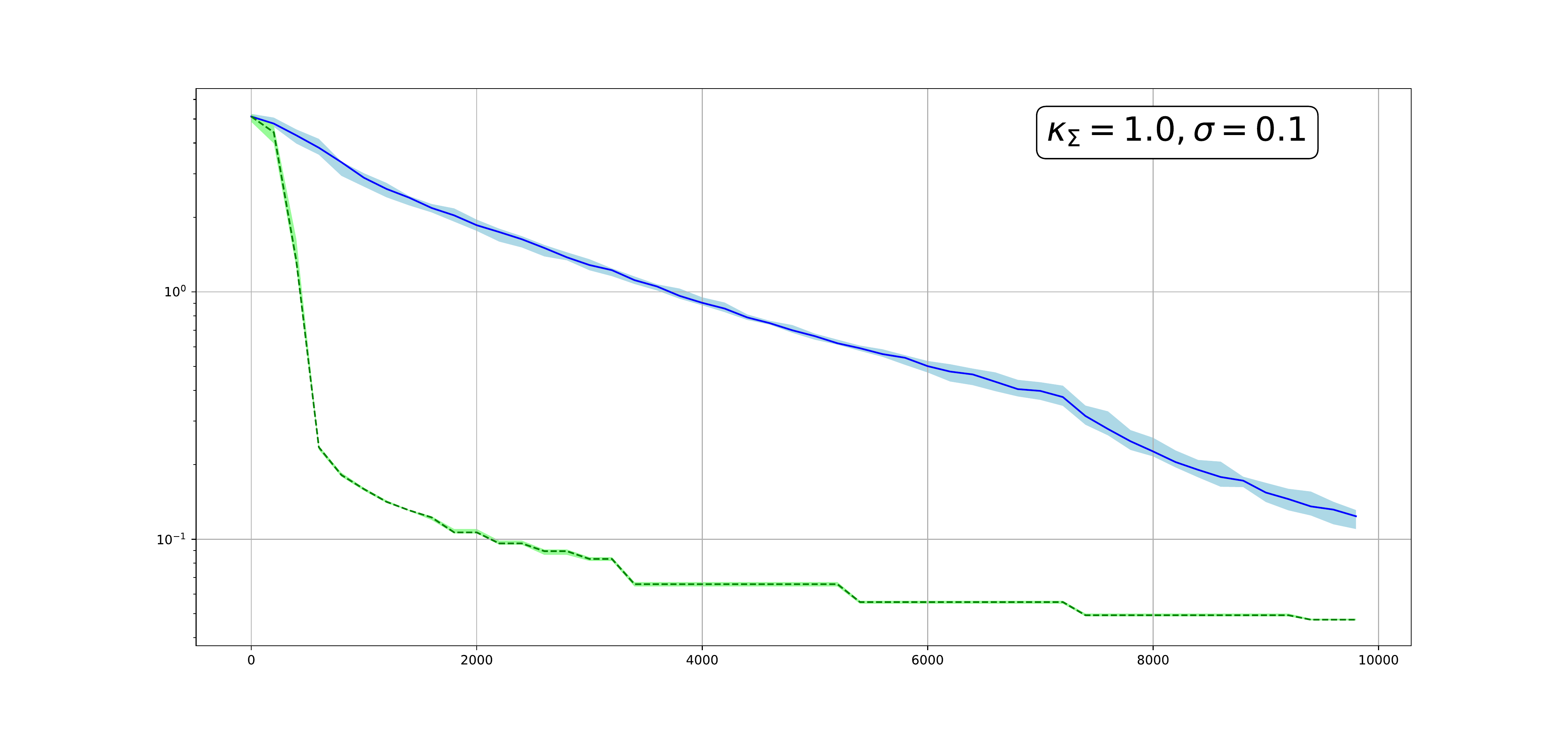}&
\includegraphics[width=0.45\textwidth]{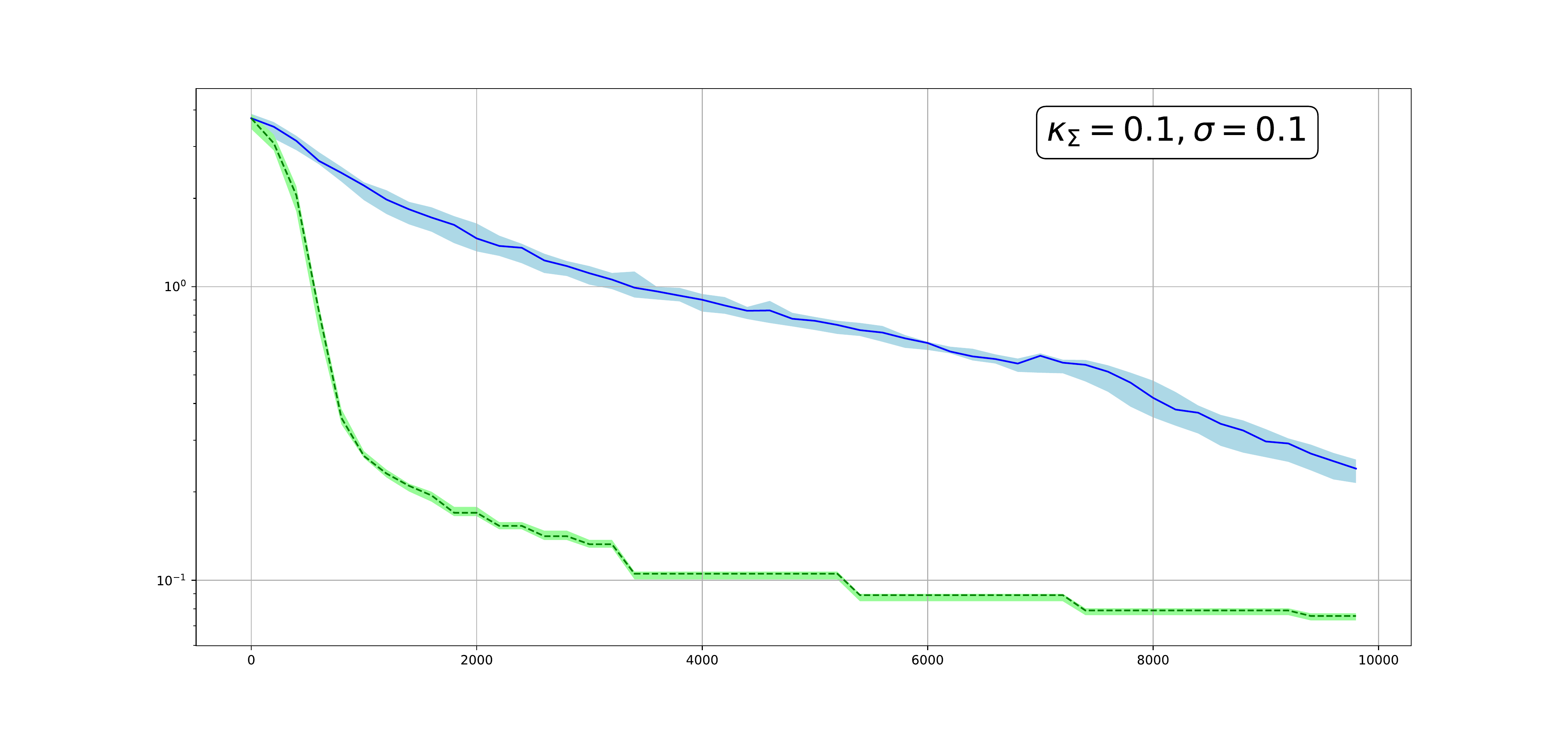}\end{tabular}
\vspace*{-0.5cm}\caption{Comparison of SMD-SR (solid line) and Lasso by CDA (dashed line) in the Gaussian setting; $(n,s) = (50\,000, 50)$.
}
\label{fig:normal.50000.lasso}
\end{figure}
\end{center}

In the first series of simulations, noises~$(\xi_i)$ are standard Gaussian, and
regressors $(\phi_i)$ are normally distributed with zero mean and covariance matrix $\Sigma$. The results for the first series are presented in Figures~\ref{fig:normal.50000} and \ref{fig:normal.50000.lasso}.
Plots in Figure~\ref{fig:normal.50000} illustrate the improvement by the SMD-SR procedure over the plain SMD algorithm in the considered settings.
The acceleration of the initial error convergence is clearly seen on the plots for~$\sigma\!=\!0.001$.

Results of a comparison with the CDA Lasso implementation of in the case of~$\sigma=0.1$ are given in Figure~\ref{fig:normal.50000.lasso}.
Because of the memory limitations of the CDA, we present the results of simulations for ~$(n,s) = (50\,000, 50)$ and $N\leq 10\,000$.
The CDA is restarted for different sizes of the observation sample, each time the number of iterations of the algorithm is limited to~$30\,000$.
{While Lasso estimate outperforms the SMD-SR for smaller observation samples, the statistical performance of the proposed algorithm appears to be competitive for large $N$.}

\begin{figure}[!hptb]
\vspace*{-0cm}
\begin{tabular}{ll}
\includegraphics[width=0.45\textwidth]{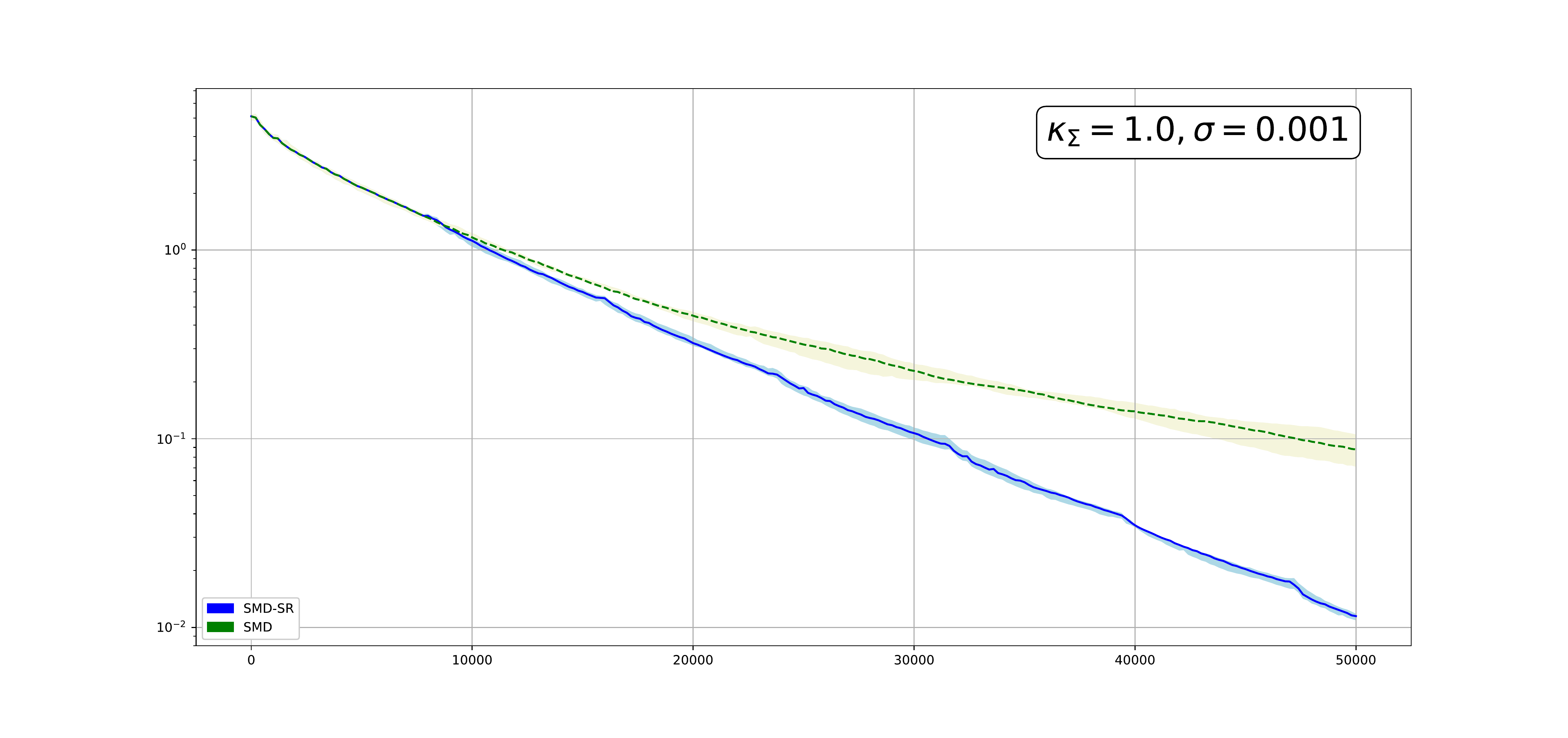}&
\includegraphics[width=0.45\textwidth]{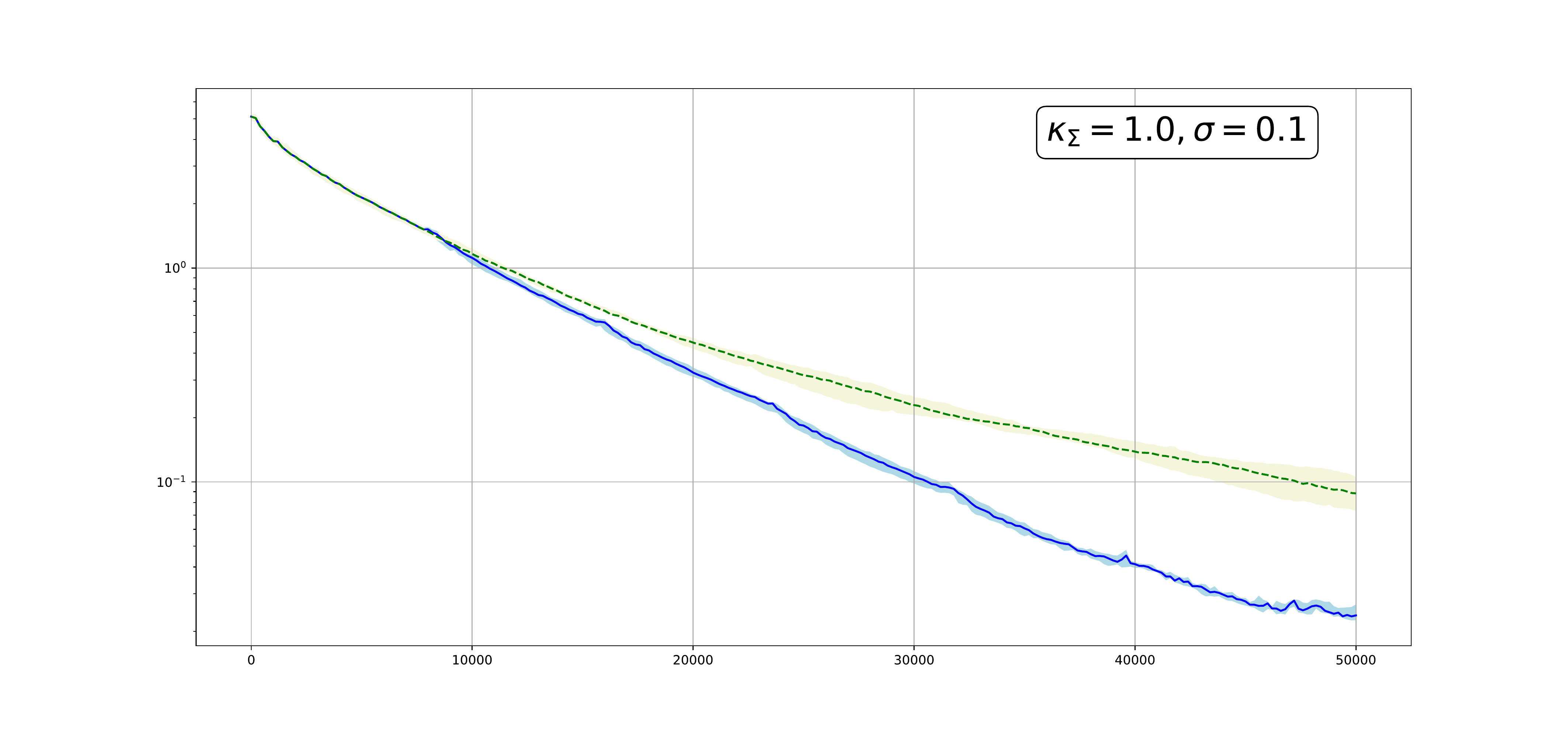}\\
\includegraphics[width=0.45\textwidth]{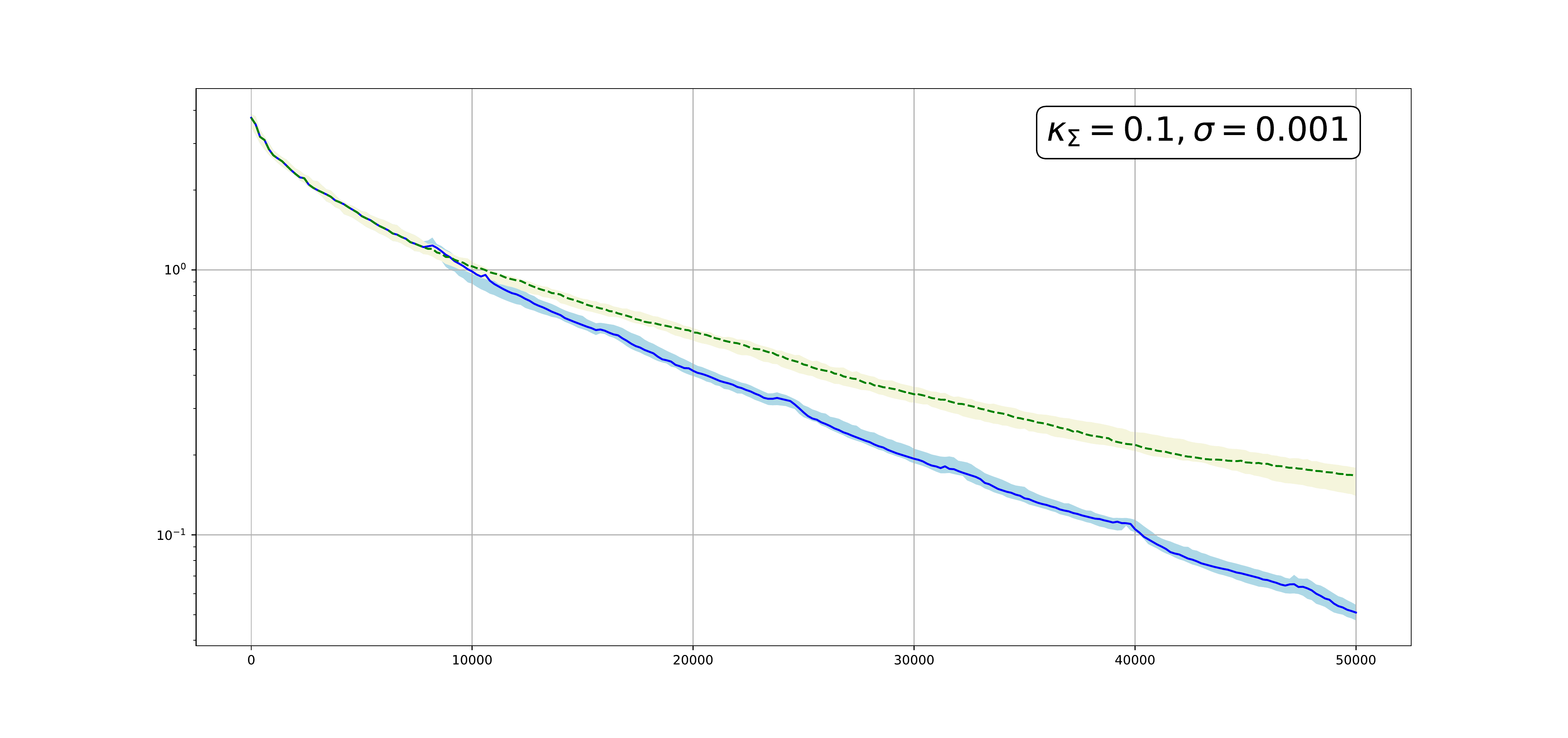}&
\includegraphics[width=0.45\textwidth]{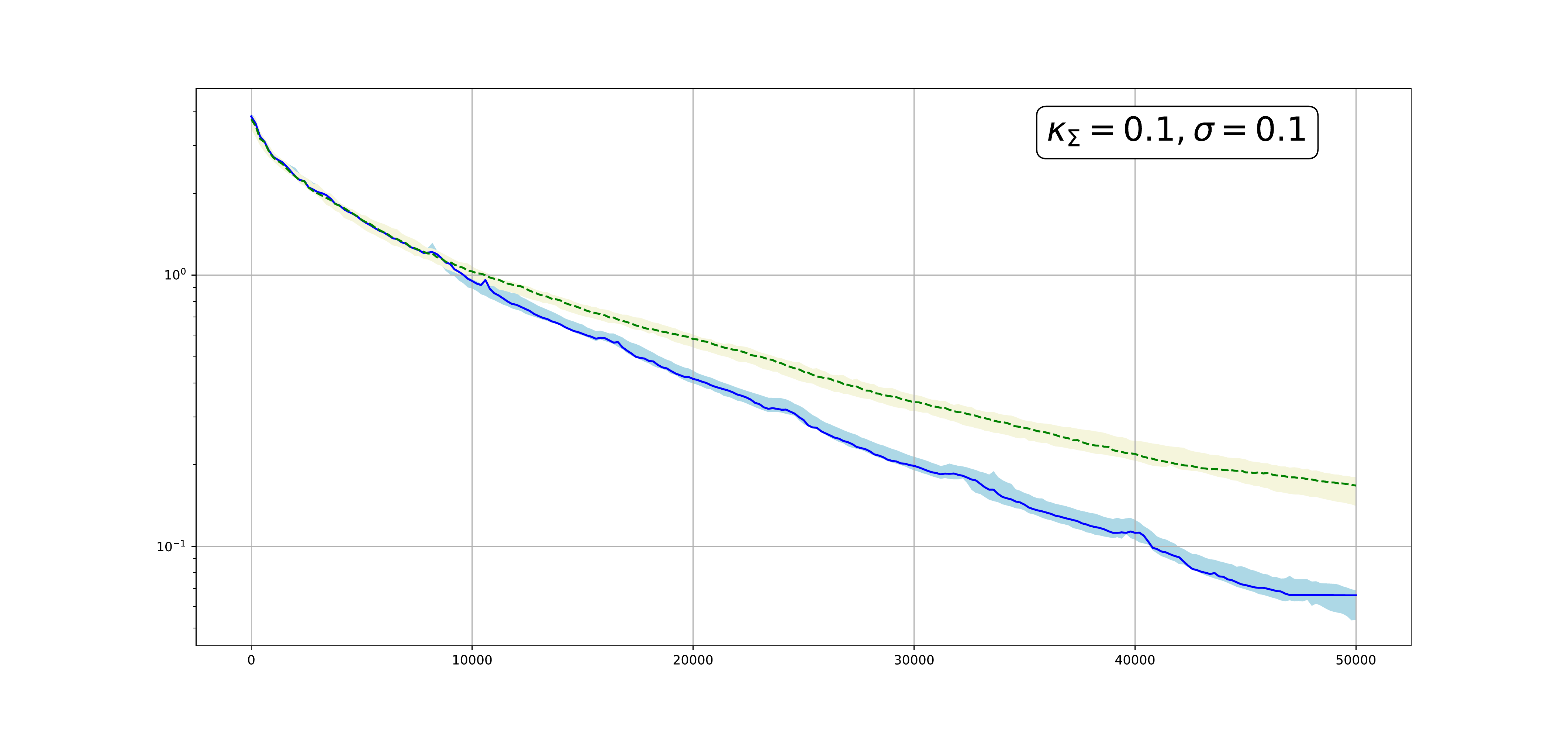}
\end{tabular}
\vspace*{-0.5cm}\caption{Comparison of SMD-SR  (solid line) and SMD  (dashed line) in the case of Student $t_4$ noise distribution;~$(n,s) = (100000, 50)$.
}
\label{fig:t4.50000}
\end{figure}

Similar results were obtained in the experiments with other types of distributions of~$\phi_i$ and~$\xi_i$.
For instance, in Figure~\ref{fig:t4.50000} 
we present the results of simulation utilizing~Student's $t_4$-distribution (i.e., multivariate Student distribution with~$4$ degrees of freedom, cf., e.g.,  \cite{kotz2004multivariate}) of noises and regressors.

\section*{Acknowledgment}
This work was supported by Multidisciplinary Institute in Artificial intelligence MIAI {@} Grenoble Alpes (ANR-19-P3IA-0003).
\appendix
\section{Proofs}

\subsection{Proof of Proposition~\ref{pr:myprop2}} We start with a technical result on the SMD algorithm which we formulate in a more general setting of composite minimization. Specifically, assume that we aim at solving the problem
	\be
	\min\limits_{x\in X} \left[f(x) =\bE\{G(x,\omega)\}+h(x)\right],
	\ee{prob1c} where $X$ and $G$ are as in Section \ref{sec:probs} and $h$ is convex and continuous.
We consider a more general {\em composite proximal mapping} \cite{nesterov2013gradient,nesterov2013first} for $\zeta\in E,\,x,x_0\in X$, and $\beta>0$ we define
	\be
	\Prox_{\beta}(\zeta,x;x_0)&:=&\argmin_{z\in X}\big\{\langle  \zeta,z\rangle +h(z)+\beta V_{x_0}(x,z)\big\}\nn
 &=&\argmin_{z\in X}\big\{\langle  \zeta-\beta \nabla\vartheta(x-x_0),z\rangle +h(z)+\beta \vartheta(z-x_0)\big\}
	\ee{cprox}
and consider for $i=1,2,\dots$ Stochastic Mirror Descent recursion \rf{eq:md1}. Same as before, the approximate solution after $N$ iterations of the algorithm is defined as weighted average of $x_i$'s according to \rf{eq:asol}. Obviously, to come back to the situation of Section \ref{sec:pstat} it suffices to put $h(x)\equiv 0$. To alleviate notation we denote $V(x,z)=V_{x_0}(x,z)$; we also denote
	\[ \zeta_i=\nabla G(x_{i-1},\omega_i)-\nabla g(x_{i-1})\]
	and
	\be
	\varepsilon(x^N,z)=\sum_{i=1}^{N}\beta^{-1}_{i-1}[\la \nabla g(x_{i-1}),x_{i}-z\ra+ h(x_{i})-h(z)]+\half V({x_{i-1}},x_i),
	\ee{epsdef}
	with $x^N=(x_0,\dots, x_N)$. In the sequel we use the following well known result which we prove below for the sake of completeness.
	\begin{proposition}\label{pr:muprop}
In the situation of this section, let $\beta_i\ge 2\L$ for all $i=0,1,...$, and let  $\widehat{x}_N$ be defined in \rf{eq:asol}, where $x_i$
		are iterations \rf{eq:md1}. Then for any $z\in X$ we have
		\be
		\left[\sum_{i=1}^{N}\beta^{-1}_{i-1}\right][f(\widehat{x}_N)-f(z)]&\le&\sum_{i=1}^{N}\beta^{-1}_{i-1}[ f(x_i)-f(z)]\leq
		\varepsilon(x^N,z)
		\nn
		&\leq & V({x_0},z)-V({x_N},z)+\sum_{i=1}^{N}\Big[
		{\la\zeta_{i},z-x_{i-1}\ra\over \beta_{i-1}} +{\|\zeta_{i}\|_*^2\over \beta_{i-1}^2}\Big]
		\label{simple0}\\
		&\leq & 2V({x_0},z)+\sum_{i=1}^{N}\Big[
		{\la\zeta_{i},z_{i-1}-x_{i-1}\ra \over \beta_{i-1}} +{ 3\over 2}{\|\zeta_{i}\|_*^2\over \beta_{i-1}^2}\Big],
		\ee{simple1}
		where $z_i$ is a random vector with values in $X$ depending only on $x_0,\zeta_1,\dots,\zeta_i$.
	\end{proposition}
\paragraph{Proof of Proposition~\ref{pr:muprop}.}
{\bf 1$^o$.} Let $ x_0, \dots, x_N $ be some points of $X$; let
		\[
		\varepsilon_{i+1}(z):=\la \nabla g(x_i),x_{i+1}-z\ra +\la h'(x_{i+1}),x_{i+1}-z\ra+\L V({x_i},x_{i+1})
		\]
(here $h'(x)$ stands for a subgradient of $h$ at $x$).
Note that  $V({x},z)\geq \half \|x-z\|^2$ due to the strong convexity of $V(x,\cdot)$. Thus, by  convexity of $g$ and $ h$ and the Lipschitz continuity of $\nabla g$ we get for any $z\in X$
		\bse
		f(x_{i+1})-f(z)&=&[ g(x_{i+1})- g(z)]+[ h(x_{i+1})- h(z)]\nn
		&=&[ g(x_{i+1})- g(x_{i})]
		+[ g(x_{i})- g(z)]+[ h(x_{i+1})- h(z)]\nn
		&\leq& [\la\nabla g(x_i),x_{i+1}-x_i\ra+\L V({x_i},x_{i+1})]+\la \nabla g(x_i),x_{i}-z\ra
		+ h(x_{i+1})- h(z)\nn
		&\leq &\la \nabla g(x_i),x_{i+1}-z\ra+\la h'(x_{i+1}),x_{i+1}-z\ra+\L V({x_i},x_{i+1})=\varepsilon_{i+1}(z);
		\ese 
i.e., the following inequality holds  for any $z\in X$:
		\be
		f(x_{i+1})-f(z)\leq \varepsilon_{i+1}(z).
		\ee{oncelem0}
{\bf 2$^o$.}
Let us first prove  inequality \rf{simple0}. 
	The optimality condition for  $x_{i+1}$ in \rf{cprox}  implies (cf. Lemma A.1 of \cite{nesterov2013first}) that there is $h'(x_{i+1})\in \partial h(x_{i+1})$ such that
	\[
	\la \nabla G(x_{i},\omega_{i+1})+h'(x_{i+1})+\beta_i\la [\nabla\vartheta(x_{i+1})-\nabla\vartheta(x_i)],z-x_{i+1}\ra\ge 0,\;\;\forall \; z\in X,
	\]
	or, equivalently,
	\bse
	\lefteqn{\la \nabla G(x_{i},\omega_{i+1})+h'(x_{i+1}),x_{i+1}-z\ra \leq \beta_i\la \nabla\vartheta(x_{i+1})-\nabla\vartheta(x_i),z-x_{i+1}\ra}\\&=&
	\beta_i\la \nabla V_{x_{i+1}}(x_i,x_{i+1}),z-x_{i+1}\ra=\beta_i[V({x_i},z)-V({x_{i+1}},z)-V({x_i},x_{i+1})],\;\;\forall \; z\in X
	\ese
where the concluding equality follows from the following remarkable identity (see, for instance, \cite{chen1993convergence}): for any $u, u'$ and $w\in X$
	\[
	\la \nabla_{u'} V(u,u'),w-u'\ra=V(u,w)-V(u',w)-V(u,u').
	\]
	This results in
	\be
	\la \nabla g (x_i),x_{i+1}-z\ra+\la h'(x_{i+1}),x_{i+1}-z\ra
	&\leq&
	\beta_i[V({x_i},z)-V({x_{i+1}},z)-V({x_i},x_{i+1})]
	\nn
	&&-\la \zeta_{i+1},x_{i+1}-z\ra.
	\ee{eq:BT}
	It follows from \rf{oncelem0} and  condition $\beta_i\geq 2\L$  that
	\bse
	f(x_{i+1})-f(z)\leq \varepsilon_{i+1}(z)\leq
	\la\nabla g(x_i),x_{i+1}-z\ra+\la h'(x_{i+1}),x_{i+1}-z\ra+{\beta_i\over 2}V({x_i},x_{i+1}).
	\ese
	Together with
	\eqref{eq:BT}, this inequality implies
	\bse
	\varepsilon_{i+1}(z)\leq \beta_i[V({x_i},z)-V({x_{i+1}},z)-\half V({x_i},x_{i+1})]-\la\zeta_{i+1},x_{i+1}-z\ra.
	\ese
	On the other hand, due to the strong convexity of $V(x,\cdot)$ we have
	\bse
	\la\zeta_{i+1},z-x_{i+1}\ra-{\beta_i\over 2}V({x_i},x_{i+1})&=&\la\zeta_{i+1},z-x_{i}\ra+\la\zeta_{i+1},x_{i}-x_{i+1}\ra-{\beta_i\over 2}V({x_i},x_{i+1})\nn
	&\le &\la\zeta_{i+1},z-x_{i}\ra+{\|\zeta_{i+1}\|^2_*\over \beta_i}.
	\ese
	Combining these inequalities, we obtain
	\be{}\quad\quad\quad\quad
	f(x_{i+1})-f(z)\leq\varepsilon_{i+1}(z)\leq
	\beta_i[V({x_i},z)-V({x_{i+1}},z)]-\la\zeta_{i+1},x_{i}-z\ra+{\|\zeta_{i+1}\|^2_*\over \beta_i} \ \quad
	\ee{almost}
	for all $z\in X$. Dividing \rf{almost}
	by $ \beta_i $ and taking the sum over $ i $ from $0$ to $ N-1 $  we obtain \rf{simple0}.
	\\{\bf 3$^o$.}
	We now prove the bound \rf{simple1}. Applying Lemma~6.1 of \cite{nemirovski2009robust} with $ z_0 = x_0 $ we get
	\be
	\forall z\in X,\quad\quad\sum_{i=1}^N\beta_{i-1}^{-1} \la\zeta_{i},z-z_{i-1}\ra\le V({x_0},z)+\half \sum_{i=1}^N \beta_{i-1}^{-2}\|\zeta_i\|_*^2,
	\ee{lemma61}
	where $z_i=\argmin_{z\in X}\big\{-\beta^{-1}_{i-1}\la\zeta_i,z\ra  + V({z_{i-1}},z)\big\}$ depend only on $z_0,\zeta_1,\dots,\zeta_{i}$. Further,
	\bse
	\sum_{i=1}^{N}\beta^{-1}_{i-1}\la\zeta_{i},z-x_{i-1}\ra&=&\sum_{i=1}^{N}\beta^{-1}_{i-1}[\la\zeta_{i},z_{i-1}-x_{i-1}\ra+
\la\zeta_{i},z-z_{i-1}\ra]
	\\&\leq& V({x_0},z)+\sum_{i=1}^N\beta^{-1}_{i-1} \la\zeta_{i},z_{i-1}-x_{i-1}\ra+\half\beta^{-2}_{i-1} \|\zeta_i\|_*^2.
	\ese
	Combining this inequality with \rf{simple0} we arrive at
	\rf{simple1}.
	\qed
\paragraph{Proof of Proposition \ref{pr:myprop2}.}
Note that, by definition, $\nu\geq \L$ and $\varkappa\geq 1$, thus, Proposition \ref{pr:muprop} can be applied to the corresponding SMD recursion. When applying recursively bound \rf{simple0} of the proposition with $z=x_*$ and $h(x)\equiv 0$ we conclude that $\bE\{V_{x_0}(x_i,x_*)\}$ is finite along with $\bE\{\|x_i-x_*\|^2\}$, and so $\bE\{\la \zeta_{i+1},x_i-x_*\ra\}=0$. Thus, after taking expectation we obtain
\bse
\sum_{i=1}^m [\bE\{g(x_i)\}-g_*]&\leq& \beta \bE\{V_{x_0}(x_0,x_*)-V_{x_0}(x_m,x_*)\}
+{
\beta^{-1} }\sum_{i=1}^m  \bE\{\|\zeta_i\|_*^2\}\nn
&\leq &\bE\{V_{x_0}(x_0,x_*)-V_{x_0}(x_m,x_*)\}\\&&+{
\beta^{-1} }\sum_{i=1}^m  \big(\varkappa\nu[\bE\{g(x_{i-1})-\la \nabla g(x_*),x_{i-1}-x_*\ra\}-g_*]+\varkappa'\varsigma_*^2\big),
\ese
which, thanks to convexity of $g$, leads to
\bse
\lefteqn{\left[1-{\varkappa\nu\over \beta}
\right]\sum_{i=1}^m [\bE\{g(x_{i})\}-g_*]+\beta \bE\{V_{x_0}(x_m,x_*)\}}\\&\leq&
\beta\bE\{V_{x_0}(x_0,x_*)\}+{\varkappa\nu\over \beta}[\bE\{g(x_{0})-\la \nabla g(x_*),x_{0}-x_*\ra\}-g_*]
+{m \varkappa'\varsigma_*^2\over \beta}.
\ese
Because,  due to convexity of $g$,
$
g(\wh x_{m })\leq {1\over m }\sum_{i=1}^m g(x_{i})
$ and
\[\bE\{g(x_{0})-\la \nabla g(x_*),x_{0}-x_*\ra\}-g_*\leq \half{\nu}\bE\{\|x_0-x_*\|^2\}\leq\half{\nu}R^2
\]
we conclude that when $\beta \geq 2\varkappa\nu$
\bse
\bE\{g(\wh x_{m })\}-g_*
\leq {2R^2 \over m}\left({\Theta\beta}+{\varkappa\nu^2\over 2\beta}\right)+
{2\varkappa'\varsigma_*^2\over \beta}
\ese
which is \rf{bouprop1}. \qed

\subsection{Proof of Theorem \ref{cor:mycor01}}
We start with the following straightforward result:
\begin{lemma}\label{lem:t001}
Let $x_*\in X\subset E$ be $s$-sparse,  $x\in X$, and let $x_s=\sparse(x)$---an optimal solution to \rf{sparseapp}. We have
\be
\|x_s-x_*\|\leq \sqrt{2s}\|x_s-x_*\|_2\leq 2\sqrt{2 s}\|x-x_*\|_2.
\ee{sparb0}
\end{lemma}
\paragraph{Proof.} Indeed, we have
\[
\|x_s-x_*\|_2\leq \|x_s-x\|_2+\|x-x_*\|_2\leq 2\|x-x_*\|_2
\]
(recall that $x_*$ is $s$-sparse). Because $x_s-x_*$ is $2s$-sparse we have by Assumption {\bf S2}
$$
\|x_s-x_*\|\leq \sqrt{2s}\|x_s-x_*\|_2\leq 2\sqrt{2s}\|x-x_*\|_2.\eqno{\mbox{\qed}}
$$
Proof of the theorem relies upon the following characterization of the properties of approximate solutions $y_k,\,x_k,\,x'_k$ and $y'_k$.
\begin{proposition}\label{pr:myprop03}
Under the premise of Theorem \ref{cor:mycor01},
\item{(i)} after $k$ preliminary stages of the
algorithm one has
\be\label{mp3a}
\bE\{\|y_k-x_*\|^2\}&\leq&2s\bE\{\|y_k-x_*\|^2_2\}\leq 2^{-k}R^2+32{\varsigma_*^2\bar s\varkappa'\over \lowkap\nu\varkappa},\\[5pt]
\bE\{g(\wh{x}_{m_0}(y_{k-1},\beta))\}-g_*&\leq& 2^{-k-4}{\lowkap R^2_{0}\over \bar s}+{2\varkappa'\varsigma_*^2\over \varkappa\nu}.
\ee{mp3b}
In particular, upon completion of $K=\overline K$ preliminary stages approximate solutions $\wh x^{(1)}$ and $\wh y^{(1)} $  satisfy
\be\label{mp30a}
\bE\{\|\wh y^{(1)}-x_*\|^2\}&\leq&2s\bE\{\|\wh y^{(1)}-x_*\|^2_2\}\leq 64{\varsigma_*^2\bar s\varkappa'\over \lowkap\nu\varkappa},\\[5pt]
\bE\{g(\wh{x}^{(1)})\}-g_*&\leq& {4\varkappa'\varsigma_*^2\over \varkappa\nu}.
\ee{mp30b}
\item{(ii)} Suppose that at least one asymptotic stage is complete. Let $r_k^2=2^{-k}r_0^2$ where $r_0^2{=} 64{\varsigma_*^2\bar s\varkappa'\over \lowkap\nu\varkappa}$. Then after $k$ stages of the asymptotic phase one has
\be\label{ap3a}
\bE\{\|y'_k-x_*\|^2\}&\leq&2s\bE\{\|y'_k-x_*\|^2_2\}\leq r_k^2=2^{-k}r_0^2,\\[5pt]
\bE\{g(\wh{x}_{m_k}(y'_{k-1},\beta))\}-g_*&\leq& {4\varsigma^2_*\varkappa'\over \beta_k}\leq  2^{-k+2}{\varsigma^2_*\varkappa'\over \varkappa\nu}.
\ee{ap3b}
\end{proposition}
{\bf Proof of the proposition.}
{\bf 1$^o$.} We first show that under the premise of the proposition the following relationship holds for $1\leq k\leq K$:
\be
\bE\{\|y_k-x_*\|^2\}\leq R^2_k:=
\half R_{k-1}^2+{16\varsigma_*\bar s\varkappa'\over \lowkap\nu\varkappa},\;\;R_0=R.
\ee{rec1}
Obviously, \rf{rec1} implies \rf{mp3a} for all $1\leq k\leq K$.
Observe that \rf{rec1} clearly holds for $k=1$. Let us now perform the recursive step $k-1\to k$. Indeed, bound \rf{bouprop1} of Proposition \ref{pr:myprop2} implies that after $m_0$ iterations of the SMD with the stepsize parameter satisfying \rf{betapr} and initial condition $x_0$ such that $\bE\{\|x_0-x_*\|^2\}\leq R_{k-1}$ one has
\be
\bE\{g(\wh x_{m_0 })\}-g_*&\leq&  {2 \over m_0}
\left[{2\Theta\varkappa\nu}+{\nu\over 4}\right]R^2_{k-1}+
{\varkappa'\varsigma_*^2\over \varkappa\nu}\nn&\leq&
{[8\Theta\varkappa+1]\nu\over 2m_0} R^2_{k-1}+{\varkappa'\varsigma_*^2\over \varkappa\nu}.
\ee{sk}
Note that when $m_0\geq 16\lowkap^{-1}\bar s (8\Theta\varkappa+1)\nu$ we have
\[
{8\bar s\over \lowkap}{[8\Theta\varkappa+1]\nu\over m_0}\leq \half.
\]
Therefore, when utilizing the bound \rf{sparb0} of Lemma \ref{lem:t001}
we get
\bse
\bE\{\|y_k-x_*\|^2\}&\leq& 2\bar s\bE\{\|y_k-x_*\|^2\}\leq 8\bar s\bE\{\|\wh x_{m_0}-x_*\|_2^2\}\leq {16\bar s\over \lowkap}[\bE\{g(\wh x_{m_0})\}-g_*]\\
&\leq& {16\bar s\over \lowkap}\left({[8\Theta\varkappa+1]\nu\over 2{m_0}} R^2_{k-1}+{\varkappa'\varsigma^2_*\over \varkappa\nu}\right)
\leq R_k^2:=\half R^2_{k-1}+{16\varsigma_*^2\bar s\varkappa'\over \lowkap\varkappa\nu}
\ese
which is \rf{rec1}.
Finally, when using \rf{sk} along with \rf{mp3a} we obtain
\[
\bE\{g(\wh{x}_{m_0}(y_{k-1},\beta))\}-g_*\leq {\lowkap R^2_{k-1}\over 32\bar s}+{\varkappa'\varsigma_*^2\over \varkappa\nu}\leq
 2^{-k-4}{\lowkap R^2_{0}\over \bar s}+{2\varkappa'\varsigma_*^2\over \varkappa\nu}
\]
what implies \rf{mp3b}. Now, \rf{mp30a} and \rf{mp30b} follow straightforwardly by applying \rf{mp3a} and \rf{mp3b} with $K=\overline K$.
\par\noindent {\bf 2$^o$.} Let us prove \rf{ap3a}. Recall that at the beginning of the first stage of the second phase we have
$\bE\{\|\bar y_0-x_*\|\}\leq r_0^2$. Now, let us do the recursive step, i.e., assume that \rf{ap3a} holds for some $0\leq k<K'$, and let us show that it holds for $k+1$.
Because $\Theta\geq 1$ and $\varkappa\geq 1$ we have $\beta^2_k\geq {\varkappa \nu^2\over 2\Theta}$, $k=1,...$, and, by \rf{bouprop1},
\be
\bE\{g(\wh x_{m_k}(y'_{k-1},\beta_k))\}-g_*&\leq&
{2r^2_{k-1} \over m_k}\left({\Theta\beta_k}+{\varkappa\nu^2\over 2\beta_k}\right)+
{2\varkappa'\varsigma_*^2\over \beta_k}\leq {4\Theta\beta_k r^2_{k-1} \over m_k}+{2\varkappa'\varsigma_*^2\over \beta_k}\nn
&\leq &2^{-k} {r_0^2\lowkap\over 64\bar s}+ 2^{1-k}{\varkappa'\varsigma_*^2\over \varkappa\nu} \leq 2^{-k} {r_0^2\lowkap\over 16\bar s}
\leq 2^{-k+2}{\varkappa'\varsigma_*^2\over \varkappa\nu}.
\ee{toap30b}
Observe that
\[
\bE\{\|x_{m_k}(y'_{k-1},\beta_k)-x_*\|_2^2\}\leq {2\over \lowkap} [\bE\{g(\wh x_{m_k}(y'_{k-1},\beta_k))\}-g_*]
\leq 2^{-k}{r_0^2\over 8\bar s},
\]
so that by Lemma \ref{lem:t001}
\[
\bE\{\|y'_{k}-x_*\|^2\}\leq 8s\bE\{\|x_{m_k}(y'_{k-1},\beta_k)-x_*\|_2^2\}\leq  2^{-k}r_0^2=r_k^2,
\]
and \rf{ap3a} follows. Now \rf{ap3b} is an immediate consequence of  \rf{ap3a} and \rf{toap30b}.\qed
\paragraph{Proof of the theorem.}
{\bf 1$^o$.} Let us start with the situation where no asymptotic stage takes place. Because we have assumed that $N$ is large enough so that at least one preliminary stage took place this can only happen when either
$m_0K\geq {N\over 2}$ or $m_1\geq {N\over 2}$. Due to $m_0>1$, by \rf{mp30a}  we have in the first case:
\[
\bE\{\|y_K-x_*\|^2\}\leq R_K^2:=2^{-K} R_0^2+{32\varsigma_*^2\bar s\varkappa'\over \lowkap\nu\varkappa}\leq 2^{-K+1} R_0^2\leq
 R_0^2\exp\left\{-{c N\lowkap\over \Theta\varkappa\bar s\nu}\right\}
\]
for some absolute $c>0$. Furthermore, due to \rf{mp3b} we also have in this case
\[
\bE\{g(\wh{x}_{m_0}(y_{K-1},\beta))\}-g_*\leq 2^{-K-4}{\lowkap R^2_{0}\over \bar s}+{2\varkappa'\varsigma_*^2\over \varkappa\nu}\leq 2^{-K-3}{\lowkap R^2_{0}\over \bar s}\leq
{\lowkap R^2_{0}\over \bar s}\exp\left\{-{c N\lowkap\over \Theta\varkappa\bar s\nu}\right\}.
\]
Next, $m_1\geq {N\over 2}$ implies that
\be
{\bar s\over \lowkap}\geq {c N\over \Theta\nu\varkappa}
\ee{form1=0}
for some absolute constant $c$, so that approximate solution $y_K$ at the end of the preliminary phase satisfies (cf. \rf{mp30a})
\[
\bE\{\|\wh y-x_*\|^2\}\leq C{\varsigma_*^2\bar s\varkappa'\over \lowkap\nu\varkappa}\leq C{\Theta \varkappa'\varsigma_*^2\bar s^2\over \lowkap^2N}.
\]
Same as above, using \rf{mp30a} and \rf{form1=0}
we conclude that in this case
\[
\bE\{g(\wh{x})\}-g_*\leq C{\varkappa'\varsigma_*^2\over \varkappa\nu}\leq C{\Theta \varkappa'\varsigma_*^2\bar s\over \lowkap N}.
\]
{\bf 2$^o$.} Now, let us suppose that at least one stage of the asymptotic phase was completed. Applying the bound \rf{ap3a} of Proposition \ref{pr:myprop03} we have
$\bE\{\|y'_{K}-x_*\|^2\}\leq r_0^2$. When $M< N/2$, same as above, we have
\[
\bE\{\|\wh y_N-x_*\|^2\}\leq r_0^2\leq  R_0^2\exp\left\{-{c N\lowkap\over \Theta\varkappa\bar s\nu}\right\}
\]
and
\be
\bE\{g(\wh{x}_{m_{K'}}(y_{K'-1},\beta))\}-g_*\leq \bE\{g(\wh{x})\}-g_*\leq
{\lowkap R^2_{0}\over \bar s}\exp\left\{-{c N\lowkap\over \Theta\varkappa\bar s\nu}\right\}.
\ee{new***}
When $M\geq N/2$, since $m_k\leq C\bar m_k$ where $\bar m_k=512{\bar s\Theta\nu\varkappa\over \lowkap}2^k$ we have
\[{N\over 2}\leq C\sum_{k=1}^{K'} \bar m_k\leq C2^{K'+1} \bar m_1\leq C2^{K'}{\bar s\Theta\nu\varkappa\over \lowkap}.
\] We conclude that
$2^{-K'}\leq C{\bar s\Theta\nu\varkappa\over \lowkap N}$ so that
\[
\bE\{\|\wh y_N-x_*\|^2\}=\bE\{\|\wh y_{K'}-x_*\|^2\}\leq 2^{-K'}r_0^2\leq C{\Theta\varkappa'\varsigma_*^2\bar s^2\over \lowkap^2 N}.
\]
Finally, by \rf{ap3b},
\[
\bE\{g(\wh{x}_{m_{K'}}(y_{K'-1},\beta))\}-g_*\leq 2^{-K'+2}{\varsigma^2_*\varkappa'\over \varkappa\nu}\leq C
{\varsigma^2_*\bar s\varkappa'\Theta\over \lowkap N};
\]
together with \rf{new***} this implies \rf{1finb}.\qed
\subsection{Proof of Theorem \ref{cor:reli}}
{\bf 1$^o$.} By the Chebyshev inequality,
\be
\forall \ell\;\;\; \Prob\{\|\wh x^{(\ell)}_M-x_*\|_2\geq 2\theta_M\}\leq \four;
\ee{cheb}
applying  \cite[Theorem 3.1]{minsker2015geometric} 
we conclude that
\[
 \Prob\{\|\wh x_{N,1-\epsilon}-x_*\|_2\geq 2C_\alpha \theta_M\}\leq e^{-L\psi(\alpha,{1\over 4})}
\]
where
\be
\psi(\alpha,\beta)=(1-\alpha)\ln {1-\alpha\over 1-\beta}+\alpha \ln{\alpha\over \beta}
\ee{psidef}
and $C_\alpha={1-\alpha\over \sqrt{1-2\alpha}}$. When choosing $\alpha={\sqrt{3}\over 2+\sqrt{3}}$ which corresponds to $C_\alpha=2$ we obtain $\psi(\alpha,\four)=0.1070...>0.1$ so that
\[
 \Prob\{\|\wh x_{N,1-\epsilon}-x_*\|_2\geq 4 \theta_M\}\leq \epsilon
 \]
if $L\geq 10\ln[1/\epsilon]$. When combining this result with that of Lemma \ref{lem:t001} we arrive at the theorem statement for solutions $\wh x_{N,1-\epsilon}$ and $\wh y_{N,1-\epsilon}$.
\\{\bf 2$^o$.}
The corresponding result for $\wh x'_{N,1-\epsilon}$ and its ``sparsification'' $\wh y'_{N,1-\epsilon}$ is due to the following simple statement.
\begin{proposition}\label{mymedian}
Let $0<\alpha<\half$, $| \cdot| $ be a norm on $E$, $z\in E$, and let $z_\ell,\,\ell=1,...,L$ be independent and satisfy
\[
 \Prob\{| z_\ell-z| \geq \delta\}\leq \beta
\]
for some $\delta>0$ and $\beta<\alpha$. Then for $\wh z$,
\be
\wh z\in \Argmin_{u\in \{z_1,...,z_L\}} \sum_{\ell=1}^L | u-z_\ell| ,
\ee{rfmymed}
it holds
\[
 \Prob\{| \wh z-z| \geq C'_\alpha \delta\}\leq e^{-L\psi(\alpha,\beta)}
\]
with $C'_\alpha= {2+\alpha\over 1-2\alpha}$.
\end{proposition}
{\bf Proof.}
W.l.o.g. we may put $\delta=1$ and $z=0$.
Proof of the proposition follows that of \cite[Theorem 3.1]{minsker2015geometric} with Lemma 2.1 of \cite{minsker2015geometric} replaced with the following result.
\begin{lemma}\label{smallem}
Let $z_1,...,z_L\in E$, and let $\wh z$ be an optimal solution to \rf{rfmymed}. Let $0<\alpha<\half$, and let $| \wh z| \geq C'_\alpha$. Then there exists a subset $I$ of $ \{1,...,L\}$ of cardinality $\card I>\alpha L$ such that for all $\ell\in I$ $| z_\ell| >1$.
\end{lemma}
{\bf Proof of the lemma.}  Let us assume that $| z_\ell| \leq 1, \,\ell=1,...,\bar L$ for $\bar L\geq (1-\alpha)L$. Then
\[\begin{array}{rcl}
\sum_{\ell=1}^L| z_\ell-\wh z| &=&\sum_{\ell\leq \bar L} | z_\ell-\wh z| +\sum_{\ell> \bar L} | z_\ell-\wh z|
\geq \bar L(C_\alpha-1)+\sum_{\ell> \bar L} [| z_\ell| -C_\alpha]\\
&\geq& \sum_{\ell\leq \bar L}| z_\ell| +\bar L(C_\alpha-2)+\sum_{\ell> \bar L} | z_\ell| -(L-\bar L)C_\alpha
\\&\geq& \sum_{\ell=1}^L| z_\ell| +
\bar L(C_\alpha-2)-(L-\bar L)C_\alpha\\
&\geq&\sum_{\ell=1}^L| z_\ell-z_1| +\bar L(2C_\alpha-2)-LC_\alpha+L-1> \sum_{\ell=2}^L| z_\ell-z_1|
\end{array}
\]
for $\bar L > {LC_\alpha+L-1\over 2(C_\alpha-1)}$. We conclude that $1-\alpha\leq {C_\alpha+1\over 2(C_\alpha-1)}$, same as
$C_\alpha\leq  {2+\alpha\over 1-2\alpha}$. \qed\par
For instance, when choosing $\alpha=1/6$ with $C_\alpha=13/4$, and $\beta$ such that $C_\alpha/\sqrt{\beta}=10$ we obtain $\psi(\alpha,\beta)=0.0171...$ so that for $L=\lceil 58.46\ln[1/\epsilon]\rceil$ we have $L\psi(\alpha,\beta)\geq \ln[1/\epsilon]$. Because
\[
\Prob\left\{\|\wh x_M^{(\ell)}-x_*\|_2\geq {\theta_M\over \sqrt{\beta}}\right\}\leq \beta,\;\;\ell=1,...,L,
\]
by Lemma \ref{smallem} we conclude that
\[
\Prob\left\{\|\wh x'_{1-\epsilon,N}-x_*\|_2\geq 10\theta_M\right\}\leq \epsilon,
\]
implying statement of  the theorem for $\wh x'_{1-\epsilon,N}$ and $\wh y'_{1-\epsilon,N}$.
\\{\bf 3$^o$.} The proof of the claim for solutions $\wh x''_{1-\epsilon,N}$ and $\wh y''_{1-\epsilon,N}$ follows the lines of that of \cite[Theorem 4]{hsu2014heavy}. We reproduce it here (with improved parameters of the procedure) to meet the needs of the proof of Theorem \ref{cor:reli2}.
\par Let us denote $I(\tau_M)$ the subset of $\{1,...,L\}\cup\emptyset$ such that $g(\wh{x}^{(i)}_M)-g_*\leq 2\tau_M$ and thus $\|\wh{x}^{(i)}_M-x_*\|_2\leq 2\theta_M$ for $i\in I(\tau_M)$. Assuming the latter set is nonempty
 we have for all $i,j\in I(\tau_M)$
$\|\wh{x}^{(i)}_M-\wh{x}^{(j)}_M\|_2\leq 4\theta_M.$
On the other hand, using \rf{cheb} and independence of $\wh{x}^{(i)}_M$ we conclude that (cf. e.g., \cite[Lemma 23]{lerasle2011robust})
\bse
\Prob\left\{|I|\geq  \rceil L/2\lceil\right\}\geq \Prob\left\{B(L,\four)\geq\rceil L/2 \lceil\right\}
\geq 1- \exp\left\{-L\psi\left({\rfloor L/2\lfloor\over L},{1\over 4}\right)\right\}
\ese
where $\rfloor a\lfloor=\lceil a\rceil-1$ is the largest integer strictly less than $a$,
$B(N,p)$ is a $(N,p)$-binomial random variable and $\psi(\cdot,\cdot) $ is as in \rf{psidef}. When $\varepsilon\leq \four$ and $L= \lceil 12.05\ln[ 1/\varepsilon]\rceil\geq 16$ we have
\[
\Prob\{|I|\geq\rceil   L/2\lceil\}\geq 1- e^{-L\psi({7\over 16},{1\over 4})}\geq 1-e^{-0.083 L}\geq 1-\varepsilon.
\]
Therefore, if we denote $\overline \Omega_\varepsilon$ a subset of $\Omega^N$ such that $|I(\tau_M)|> L/2$ for  $\omega^N\in \overline\Omega_\epsilon$ we have
$P\{\overline\Omega_\varepsilon\}\geq 1-\varepsilon.$ Let now $\omega^N\in \overline\Omega_\varepsilon$ be fixed. Observe that the optimal value $\wh r=r^{\wh i} _{\rceil L/2\lceil}$ of \rf{3rdmed} satisfies $\wh r\leq 4\theta_M$, and that among $\rceil L/2\lceil$ closest to $\wh x''_{N,1-\epsilon}$ points there is at least one, let it be $\wh{x}^{(\bar i)}_M$ satisfying $g(\wh{x}^{(\bar i)}_M)-g_*\leq 2\tau_M$ and $\|\wh{x}^{(\bar i)}_M-x_*\|_2\leq 2\theta_M$. We conclude that whenever $\omega^N\in \overline\Omega$ one has
\[\|\wh x''_{N,1-\epsilon}-x_*\|_2\leq \|\wh x''_{N,1-\epsilon}-\wh{x}^{(\bar i)}_M\|_2+\|\wh{x}^{(\bar i)}_M-x_*\|_2\leq4\theta_M+2\theta_M\leq 6 \theta_M,
\]
implying that
\[
\Prob\{\|\wh x''_{N,1-\epsilon}-x_*\|_2\geq 6\theta_M\}\leq \varepsilon
\]
whenever $L\geq \lceil 12.05\ln[ 1/\varepsilon]\rceil$. \qed
\subsection{Proof of Theorem \ref{cor:reli2}}
The proof of the theorem relies on the following statement which may be of independent interest.
\begin{proposition}\label{pr:compar}
Let $U:\;[0,1]\times \Omega\to \bR$ be continuously differentiable and such that $u(t)=\bE\{U(t,\omega)\}$ is finite for all $t\in[0,1]$, convex and differentiable with Lipschitz-continuous gradient:
\[
|u'(t')-u'(t)|_*\leq \M|t-t'|,\qquad \forall \,t,t'\in [0,1].
\]
In the situation in question, let $\varepsilon\in(0,\four]$, $J\geq \Big\lceil 7\ln [2/\varepsilon]\Big\rceil$, and $t_i={2i-1\over 2m}$,  $i=1,...,m$. Consider the estimate
  \[
  \wh v=\med_j [\wh v^j],\;\;\wh v^j= {1\over m}\sum_{i=1}^m U'(t_i,\omega^j_i)\;\;j=1,...,J
  \]
of the difference $v=u(1)-u(0)$ using $M=mJ$ independent realizations $\omega^j_i$, $i=1,...,m,\,j=1,...,L$. Then
\be \Prob\{|\wh v-v|\geq \rho \}\leq {\varepsilon}
\ee{robustestp}
where
\[
\rho ={1\over 4m}\left[\sqrt{2\M(u(1)-u_*)}+\sqrt{2\M(u(0)-u_*)}\right]
+{2\over m}\sqrt{\sum_{i=1}^m \bE\left\{[\zeta^1(t_i)]^2\right\}},
\]
(here and below, $\zeta^j(t_i)=U'(t_i,\omega^j_i)-u'(t_i)$ and $u_*=\min_{0\leq t\leq 1} u(t)$).
\par In particular, if for $\mu\geq \M$
\be
\bE\{[\zeta^1(t)]^2\}\leq \mu (u(t)-u_*)+\varsigma^2
\ee{notS2}
then
\be \Prob\{|\wh v-v|\geq \bar\rho \}\leq {\varepsilon}
\ee{robustest}
where
\[
\bar\rho=2\sqrt{\mu\over m}
\left[\sqrt{u(1)-u_*}+\sqrt{u(0)-u_*}\right]+{2\varsigma\over \sqrt{m}}.
\]\end{proposition}
We postpone the proof of the proposition to the end of this section.

\noindent{\bf 1$^o$.} Let $\omega^N\in \overline \Omega_{\epsilon/2}$ defined as in 3$^o$ of the proof of Theorem \ref{cor:reli}; we choose $L\geq  \lceil 12.05\ln[ 2/\varepsilon]\rceil$ so that $\Prob\{\overline \Omega_{\epsilon/2}\}\leq \epsilon/2$.  We denote $\wh{r}$ the optimal value of \rf{3rdmed}; recall that $\wh r\leq 4\theta_M$. Then for any $i,j\in \wh I$ we have
\be
\|\wh x^{(i)}_{M}-\wh x^{(j)}_{M}\|_2\leq 2\wh r\leq 8\theta_M,
\ee{rijbnd} and for some $\bar i\in \wh I$ we have
\be
 g(\wh{x}^{(\bar i)}_M)-g_*\leq2\tau_M^2
 \ee{gwhbnd}
 where $\tau_M$ and $\theta_M$ are defined in \rf{Mbound1} and \rf{Mbound2} respectively.
W.l.o.g. we can assume that $\wh{x}^{(\bar i)}_M$ is the minimizer of $g(x)$ over $\wh x^{(i)}_M$, $i\in \wh I$.
\par
Let us consider the aggregation procedure. From now on all probabilities are assumed to be computed with respect to the distribution $P^K$ of the (second) sample $\omega^K$, conditional to realization $\omega^N$ of the first sample (independent of $\omega^K$).
To alleviate notation we drop the corresponding ``conditional indices.''
\\{\bf 2$^o$.} Denote $\wh v_{ji}=\med_\ell [\wh v^\ell_{ji}]$. For $j\in \wh I$, $j\neq \bar i $ let $x(t)=\wh x^{(j)}_M+t\big(\wh x^{({\bar i})}_M-\wh x^{(j)}_M\big)$. Note that $U(t,\omega)=G(x(t),\omega)$ and $u(t)=g(x(t))$  satisfy the premise of Proposition \ref{pr:compar} with
$\M=r^2_{j\bar i} \L_2$ where $r_{j\bar i}=\| \wh x^{({\bar i})}_M-\wh x^{(j)}_M\|_2$, $\mu=\chi \L_2r^2_{j\bar i}$, and $\varsigma^2=\chi'\varsigma_*^2r^2_{j\bar i}$. When applying the proposition with  $\varepsilon={\epsilon/ L}$, $J=L'$, and $K=mL'$ we conclude that
\[
\forall j\in \wh I,\,j\neq \bar i\;\;\;\Prob\{|\wh v_{j\bar i}-v_{j\bar i}|\geq \varrho_{j\bar i}\}\leq  {\epsilon\over L},
\]
implying that
\be
\Prob\{\max_{j\in \wh I,j\neq \bar i}|\wh v_{ j\bar i}-v_{j\bar i}|\geq \varrho_{j\bar i}\}\leq  {\epsilon\over 2}
\ee{allcomp}
where
\[
\varrho_{ij}=2r_{j\bar i}\sqrt{\L_2\chi\over m}
\left[\sqrt{g(\wh x^{(i)}_M)-g_*}+\sqrt{g(\wh x^{(j)}_M)-g_*}\right]+2r_{j\bar i}\varsigma_*\sqrt{\chi'\over m}.
\]
Let now  $\Omega'_{\epsilon/2}\subset \Omega^K$ such that
for all
\[\max_{\bar i\neq j\in \wh I}|\wh v_{j\bar i}-v_{j\bar i}|\leq \varrho_{j\bar i},\;\;\forall \omega^K\in \Omega'_{\epsilon/2};
\]
by \rf{allcomp} $\Prob\{\Omega'_{\epsilon/2}\}\geq 1-\epsilon/2$.
\\{\bf 3$^o$.} Let us fix $\omega^K\in \Omega'_{\epsilon/2}$; our current objective is to show that in this case the set of admissible $\wh x^{(i)}_M$'s is nonempty---it contains $\wh{x}^{(\bar i)}_M$---and, moreover,
all admissible  $\wh{x}^{(j)}_M$'s satisfy the bound $g(\wh{x}^{(j)}_M)\leq \gamma^2(r_{\bar ij})$ 
with $\gamma(r)$ defined as in \rf{***gam}.
\par Let $\alpha,\beta,\tau>0$, and let $v(\gamma)=\gamma^2-\tau^2 -2[\alpha(\gamma+\tau)+\beta]$; then $v(\gamma)> 0$  for
$\gamma\geq \sqrt{(2\alpha+\tau)^2+4\beta}$. Indeed, $v(\cdot)$ being nondecreasing for $\gamma\geq \alpha$, it suffices to verify the inequality for $\gamma= \sqrt{(2\alpha+\tau)^2+4\beta}$. Because
\[
2\alpha+\tau+\beta/\alpha> \sqrt{(2\alpha+\tau)^2+4\beta}
\]
we have
\[
4\alpha^2+4\alpha\tau+2\beta> 2\alpha\left(\sqrt{(2\alpha+\tau)^2+4\beta}+\tau\right),
\]
and
\[
v(\gamma)=[(2\alpha+\tau)^2+4\beta]-\tau^2-2\alpha\left(\sqrt{(2\alpha+\tau)^2+4\beta}+\tau\right)-2\beta>0.
\]
Applying the above observation to $\alpha=2r_{j\bar i}\sqrt{\L_2\chi\over m}$, $\beta=2r_{j\bar i}\varsigma_* \sqrt{\chi'\over m}$, and $\tau=\tau_M$ we
conclude that
whenever $g(\wh x^{(j)}_M)-g_*\geq \gamma^2(r_{j\bar i})$
\be
v_{j\bar i}=g(\wh x^{(\bar i)}_M)-g(\wh x^{(j)}_M)\leq \tau^2_M-g(\wh x^{(j)}_M)<-2\varrho_{j\bar i}.
\ee{vjbari} Therefore, for $g(\wh x^{(j)}_M)\geq \gamma^2(r_{j\bar i})$
\[
\med_\ell [\wh v^\ell_{j\bar i}]-\rho_{\bar ij}= [\med_\ell [\wh v^\ell_{j\bar i}]-v_{j\bar i}]+v_{j\bar i}-\rho_{\bar i j}< \varrho_{j\bar i}-2\varrho_{j\bar i}-\rho_{\bar i j}<0\;\;\forall\,\omega^K\in \Omega'_{\epsilon/2}.
\]
Furthermore, for $g(\wh x^{(j)}_M)-g_*< \gamma^2(r_{j\bar i})$ we have
\[
\med_\ell [\wh v^\ell_{j\bar i}]-\rho_{\bar ij}\leq \varrho_{\bar i j}-\rho_{\bar i j}< 0\;\;\;\forall\,\omega^K\in \Omega'_{\epsilon/2},
\]
and we conclude that $\wh x^{(\bar i)}_M$ is admissible.
\par
On the other hand, whenever $g(\wh x^{(j)}_M)-g_*\geq \gamma^2(r_{j\bar i})$ we have $v_{\bar i j}>2\varrho_{\bar ij}$ (cf. \rf{vjbari}), and
\[
\med_\ell [\wh v^\ell_{\bar i j}]-\rho_{j\bar i}= [\med_\ell [\wh v^\ell_{j\bar i}]-v_{\bar i j}]+v_{\bar i j}-\rho_{\bar i j}> -\varrho_{\bar i j}+2\varrho_{\bar i j}-\rho_{\bar i j}\geq 0\;\;\forall\,\omega^K\in \Omega'_{\epsilon/2}.
\]
We conclude that $\wh x^{(j)}_M$ is not admissible if $g(\wh x^{(j)}_M)\geq  \gamma^2(r_{j\bar i})$ and $\omega^K\in \Omega'_{\epsilon/2}$.
\\{\bf 4$^o$.} Now we are done. So, assume that $[\omega^N,\omega^K]\in \overline \Omega_{\epsilon/2}\times \Omega'_{\epsilon/2}$ (what is the case with probability $\geq 1-\epsilon$). We
have $r_{ij}\leq 8\theta_M$ for $i,j\in \wh I$ by \rf{rijbnd}, and $g(x^{(\bar i)}_M)\leq \tau^2_M$ for some admissible $\bar i\in \wh I$ by \rf{gwhbnd}. In this situation, all $\wh x^{( j)}_M$ such that
$g(\wh x^{(j)}_M)-g_*\geq \gamma^2(r_{j\bar i})$, $j\in \wh I$, are not admissible, implying that the suboptimality of the selected solution $\overline x_{N+K,1-\epsilon}$ is bounded with $\gamma^2(8\theta_M)$, thus
$$\risk_{g,\epsilon}(\overline x_{N+K,1-\epsilon}|X)\leq \bar \gamma^2=\gamma^2(8\theta_M).
$$
The ``in particular'' part of the statement of the theorem can be verified by direct substitution of the corresponding values of $m$, $\theta_M$, and $\tau_M$ into the expression for  $\bar \gamma^2$.\qed
\paragraph{Proof of Proposition \ref{pr:compar}.} Let us denote
 \[\bar v=\bE\{\wh v^j\}={1\over m}\sum_{i=1}^m u'(t_i);\] we have
\be
|\wh v-v|\leq |\wh v-\bar v|+|\bar v- v|.
\ee{substituteall}
{\bf 1$^o$.} Note that
\[
\wh v^j-\bar v= {1\over m}\sum_{i=1}^m U'(t_i,\omega^j_i)-u'(t_i)={1\over m}\sum_{i=1}^m \zeta^j(t_i),
\]
and
\bse
\bE\{(\wh v^j-\bar v)^2\}
\leq {1\over m^2}\sum_{i=1}^m
\bE\{[\zeta^j(t_i)]^2\}=:\upsilon^2.
\ese
By the Chebyshev inequality, $\Prob\{ |\wh v^j-\bar v|\geq 2\upsilon\}\leq \four$, and
\bse
\Prob\{\med_j[\wh v^j]-\bar v\geq 2\upsilon\}&\leq&  \Prob\Big\{\sum_j 1\{\wh v^j-\bar v\geq 2\upsilon\}\geq {J/ 2}\Big\}\\&\leq&
\Prob\{B(J,\four)\geq J/2\}\leq e^{-J\psi(\half,\four)}\leq e^{-0.1438 J}
\ese
where 
$\psi(\cdot,\cdot)$ is defined in \rf{psidef}. Because the same bound holds for
$\Prob\{\med_j[\wh v^j]-\bar v\leq -2\upsilon\}$ we conclude that
\be
\Prob\{|\wh v-\bar v|\geq 2\upsilon\}=\Prob\{|\med_j[\wh v^j]-\bar v|\geq 2\upsilon\}\leq 2e^{-J/7}\leq \varepsilon
\ee{probmed}
for $J\geq 7\ln(2/\varepsilon)$. Furthermore, if \rf{notS2} holds we have
\bse
\bE\{(\wh v^j-\bar v)^2\}
\leq {1\over m^2}\sum_{i=1}^m[\mu (u(t_i)-g_*)+\varsigma^2]\leq {1\over 2m} [(u(1)-u*)+(u(0)-u_*)] +{\varsigma^2\over m}=:\bar\upsilon^2
\ese
implying \rf{probmed} with $\upsilon$ replaced with $\bar\upsilon$:
\be
\Prob\{|\wh v-\bar v|\geq 2\bar\upsilon\}\leq  2e^{-J/7}\leq \varepsilon
\ee{probmed1}
\\
{\bf 2$^o$.} Next, we bound the difference $\bar v- v$.
Let $s_i=i/m$, $i=0,...,m$, and $r_i=u'(s_i)-u'(s_{i-1})$. Let us show that
\[
v -{\bar v}\leq {1\over 4m}
\left[\sqrt{2\M(u(1)-u_*)}+\sqrt{2\M(u(0)-u_*)}\right].
\] Note that
\[
\delta_i=\int_{s_{i-1}}^{s_i}[u'(s)-u'(t_i)]ds\leq \four{r_i(s_i-s_{i-1})}={(4m)^{-1}r_i},
\]
so that
\[
v -{\bar v}\leq \sum_{i=1}^m \delta_i\leq (4m)^{-1}[u'(1)-u'(0)].
\]
Let now $t_*\in[0,1]$ be a minimizer of $u$ on $[0,1]$. Due to the smoothness and convexity of $u$ we have
\[
|u'(0)-u'(t_*)|^2\leq 2\M[u(0)-u_*+t_*u'(t_*)]\leq 2\M[u(0)-u_*]
\]
and
\[
|u'(1)-u'(t_*)|^2\leq 2\M[u(1)-u_*-(1-t_*)u'(t_*)]\leq 2\M[u(1)-u_*].
\]
We conclude that
\[
u'(1)-u'(0)\leq u'(1)-u'(t_*)+u'(t_*)-u'(0)\leq \sqrt{2\M[u(0)-u_*]}+\sqrt{2\M[u(1)-u_*]},
\]
and
\[
v -{\bar v}\leq (4m)^{-1}[u'(1)-u'(0)]\leq {1\over 4m}\left[\sqrt{2\M[u(0)-u_*]}+\sqrt{2\M[u(1)-u_*]}\right].
\]
The proof of the corresponding bound for ${\bar v}-v$ is completely analogous, implying that
\[
|v -{\bar v}|\leq {1\over 4m}
\left[\sqrt{2\M(u(1)-u_*)}+\sqrt{2\M(u(0)-u_*)}\right].
\]

When substituting the latter bound and the bound \rf{probmed} into \rf{substituteall} we obtain
\[
\Prob\{|\wh v-v|\geq 2\upsilon+\upsilon'\}\leq \varepsilon
\]
for $J\geq 7\ln(2/\varepsilon)$, what implies \rf{robustestp}.
When replacing \rf{probmed} with \rf{probmed1} in the above derivation we obtain \rf{robustest}.\qed
\subsection{Proofs for Section \ref{sec:matrec}}
The following statement is essentially well known:
\begin{lemma}
\label{lem:spectral.conc}
Let $\phi\in \bR^{p\times q}$ with $q\leq p$ for the sake of definiteness, be a random sub-Gaussian matrix $\phi \sim \SG(0,S)$ implying that
\be
\forall x\in \bR^{p\times q},\;\;\bE\big\{e^{\la x,\phi\ra}\big\}\leq e^{{1\over 2}\la x,S(x)\ra}.
\ee{subg}
Suppose that $S\preceq \bar s I$; then
\[
\bE\{\|\phi\|_*^2\}\le C\bar s (p+q)
\quad\text{and}\quad
\bE\{\|\phi\|_*^4\} \le C' \bar s^2 (p+q)^2
\]
where $C$ and $C'$ are absolute constants.
\end{lemma}
{\bf Proof of the lemma.}
\paragraph{1$^o$.}
Let $u\in \bR^q$ be such that $\|u\|_2=1$.
Then the random vector $\zeta=\phi u\in \bR^p$ is sub-Gaussian with~$\zeta \sim \SG(0,Q)$,
that is for any $v\in \bR^p$
\[
\bE\big\{e^{v^T\zeta}\big\}=\bE\big\{e^{v^T\phi u}\big\}=\bE\big\{e^{\la uv^T,\phi\ra }\big\}\leq e^{\half \la uv^T,S(uv^T)\ra}=e^{\half v^TQv}
\]
where $Q=Q^T\in \bR^{p\times p}$. Note that
\[
\max_{\|v\|_2 = 1} v^TQv = \max_{\|v\|_2=1} \la uv^T, S(uv^T)\ra \leq
\max_{\|w\|_2 = 1} \la w, {S}(w)\ra.
\]
Therefore, we have $Q\preceq \bar s I$, and $\Tr(Q)\leq \bar s p$.
\paragraph{2$^o$.}
Let $\Gamma=\{u\in\bR^q:\,\|u\|_2=1\}$, and let $\D_\epsilon$ be a minimal $\epsilon$-net,
w.r.t. $\|\cdot\|_2$, in $\Gamma$, and let $\N_\epsilon$ be the cardinality of $\D_\epsilon$.
We claim that
\be \left\{u^T\phi^T\phi u\leq \upsilon\;\forall u\in \D_\epsilon\right\}\;\Rightarrow \;\left\{\|\phi^T\phi\|_*\leq (1-2\epsilon)^{-1}\upsilon\,\right\}.
\ee{123claim}
Indeed, let the premise in (\ref{123claim}) hold true; $\phi^T\phi$ is symmetric, so let $\bar v\in \Gamma$ be such that $\bar v^T\phi^T\phi \bar v=\|\phi^T\phi\|_*$. There exists $u\in\D_\epsilon$ such that $\|\bar{v}-u\|_2\leq\epsilon$, whence
\[
\|\phi^T\phi\|_*=|\bar{v}^T\phi^T\phi\bar{v}|\leq 2\|\phi^T\phi\|_*\|\bar{v}-u\|_2+|u^T\phi^T\phi u|\leq 2\|\phi^T\phi\|_*\epsilon+\upsilon
\]
(note that the quadratic form $z^TQz$ is Lipschitz continuous on $\Gamma$, with constant $2\|Q\|_*$ w.r.t. $\|\cdot\|_2$), whence $\|\phi^T\phi\|_*\leq (1-2\epsilon)^{-1}\upsilon$.
\paragraph{3$^0$.} We can straightforwardly build an $\epsilon$-net $\D'$  in $\Gamma$ in such a way that the $\|\cdot\|_2$-distance between every two distinct points of the net is $>\epsilon$, so that the balls
$B_v=\{z\in\bR^p:\|z-v\|_2\leq\epsilon/2\}$ with $v\in \D'$ are mutually disjoint. Since the union of these balls belongs to $B=\{z\in\bR^q:\;\|z\|_2\leq 1+\epsilon/2\}$, we get $\Card(\D')(\epsilon/2)^q\leq (1+\epsilon/2)^q$, that is, $\N_\epsilon\leq\Card(\D')\leq (1+2/\epsilon)^q$.
\par
Now we need the following well-known result (we present its proof at the end of this section for the sake of completeness).
\begin{lemma}\label{lem:sgus}
Let $\zeta\sim \SG(0,Q)$ be a sub-Gaussian random vector in $\bR^n$, i.e.
\be
\forall t\in \bR^n\;\;\;\bE\big\{e^{t^T\zeta}\big\}\leq e^{\half t^TQt}
\ee{l.subg}
where $Q=Q^T\in \bR^{n\times n}$.
Then for all $x\geq 0$
\be
\Prob\{ \|\zeta\|_2^2\geq \Tr(Q)+ 2\sqrt{xv}+2x\bar q\}\leq e^{-x}
\ee{tdev0}
where $\bar q=\max_i \sigma_i(Q)$ is the principal eigenvalue of $Q$ and $v=\|Q\|^2_2=\sum_{i} \sigma_i^2(Q)$ is the squared Frobenius norm of $Q$. Thus, for any $\alpha>0$
\be
\Prob\{ \|\zeta\|_2^2\geq \Tr(Q)(1+\alpha^{-1})+(2+\alpha) x\bar q\}\leq e^{-x}.
\ee{tdev}
\end{lemma}
Utilizing \rf{tdev} with $\alpha=1$ we conclude that $\forall u\in \Gamma$ the random vector $\zeta=\phi u$ satisfies
\be
\Prob\{ \|\zeta\|_2^2\geq 2\bar s p+3\bar s x\}\leq e^{-x}.
\ee{tdevA}
Let us set $\epsilon=\four$; utilizing \rf{tdevA}, we conclude that the probability of violating the premise in (\ref{123claim}) with $\upsilon=2\bar s p+3\bar s x$
does not exceed $\exp\{-x+q\ln[1+2\epsilon^{-1}]\}=\exp\{-x+q\ln 9\}$, so that
\[
\Prob\left\{\|\phi^T\phi\|_*\geq 2\bar s (2p+3 x)\right\}\le \exp\{-x+q\ln 9\}.
\]
Now we are done: recall that
\bse
\bE\{\|\phi\|_*^4\}&=&\bE\{\|\phi^T\phi\|_*^2\}=2\int_{0}^\infty \Prob\{\|\phi^T\phi\|_*\geq u\}u\,du\\&\leq&
2\int_{0}^\infty u\min\Big\{\exp\Big\{{4\bar s p-u\over {6}\bar s}+q\ln 9\Big\},1\Big\}du \\&\leq& 2\int_0^{\bar s(4p+{6}q\ln 9)}udu+
2\int_{\bar s (4p+{6}q\ln 9)}^\infty u \exp\Big\{{4\bar s p-u\over {6}\bar s}+q\ln 9\Big\} du\\&\leq &
\bar s^2 (4p+{6}q\ln 9)^2+{12} \bar s^2(4p+{6}q\ln9)+72\bar s^2\leq C' \bar s^2(p+q)^2.
\ese
Similarly we  get $\bE\{\|\phi\|_*^2\}\le C\bar s (p+q)$ for an appropriate $C$.

\paragraph{4$^o$.} Let us now prove Lemma \ref{lem:sgus}.

Note that for $t<1/(2\bar s)$ and $\eta\in \bR^n$, $\eta\sim \N(0,I)$ independent of $\zeta$ we have by \rf{subg}
\bse\bE\big\{e^{ t\la\zeta,\zeta\ra}\big\}&=&\bE\left\{\bE_\eta\big\{e^{\sqrt{2t}\la \zeta,\eta\ra }\big\}\right\}=\bE_\eta\left\{\bE\big\{e^{\sqrt{2t}\la \zeta,\eta\ra}\big\}\right\}\leq \bE_\eta\big\{e^{t\la \eta,S\eta\ra}\big\}=
\bE_\eta\big\{e^{t\la \eta,D \eta\ra}\big\}\\&=&
\prod_i\bE_{\eta_i}\big\{e^{t\eta_i^2s_{i}}\big\}=\prod_i (1-2ts_i)^{-1/2}
\ese
where $D=\Diag(s_i)$ is the diagonal matrix of eigenvalues.
Recall that one has, cf. \cite[Lemma 8]{birge1998minimum},
\[-\half \ln (1-2ts_i)-ts_i \leq{t^2s_i^2\over 1-2ts_i}\leq{t^2s_i^2\over 1-2t\bar s}
\]
for $t<1/(2\bar s)$.
On the other hand, $\forall t<1/(2\bar s)$
\bse
\Prob\{\|\zeta\|_2^2-\Tr(S)\geq u\}&\leq& \bE\Big\{\exp\Big\{t\big[\|\zeta\|_2^2-\sum_i s_i-u\big]\Big\} \Big\}\\&\leq& \exp\Big\{-tu+{t^2\over 1-2t\bar s}\sum_i s_i^2\Big\}=\exp\Big\{-tu+{t^2v\over 1-2t\bar s}\Big\}.
\ese
When choosing $t={\sqrt{x}\over v+2\bar s\sqrt{x}}\,\left(<{1\over 2\bar s}\right)$ and $u=2\sqrt{xv}+2x\bar s$ we obtain
\[
\Prob\{ \|\zeta\|_2^2\geq \Tr(S)+ 2\sqrt{xv}+2x\bar s\}\leq e^{-x}\]
which is \rf{tdev0}.
Because $v\leq \Tr(S)\bar s$ the latter bound also implies \rf{tdev}.\qed


\end{document}